\documentclass{article}

\usepackage{microtype}
\usepackage{graphicx}
\usepackage{subfigure}
\usepackage{booktabs} % for professional tables
\usepackage{makecell}
\usepackage{amssymb}
\usepackage{diagbox}
\usepackage{bbding}

\usepackage{hyperref}

\usepackage[accepted]{icml2025}
\usepackage{amsmath}
\usepackage{amssymb}
\usepackage{mathtools}
\usepackage{amsthm}
\usepackage{enumitem}
\usepackage{bbm}
\usepackage{csquotes}
\usepackage{booktabs}
\usepackage{footnote}      
\makesavenoteenv{table}

\usepackage[capitalize,noabbrev]{cleveref}

\theoremstyle{plain}

\theoremstyle{definition}

\theoremstyle{remark}

\usepackage[ruled,linesnumbered]{algorithm2e}

\usepackage{color}
\usepackage{tcolorbox}
\usepackage{soul}
\tcbuselibrary{skins}
\tcbuselibrary{breakable}

\usepackage[textsize=tiny]{todonotes}
\usepackage[fixed]{fontawesome5}
\usepackage{multirow}
\usepackage{multicol}
\usepackage{caption}
\usepackage{mathrsfs}

\usepackage{colortbl}
\definecolor{my_green}{RGB}{51,102,0}
\definecolor{my_red}{RGB}{204, 0, 0}
\definecolor{my_purple}{RGB}{160, 43, 147}
\definecolor{my_blue}{RGB}{15, 158, 213}
\definecolor{my_orange}{RGB}{255, 127, 0}
\newcommand{\ours}{{GLIDER}}

\icmltitlerunning{Divide and Conquer: Grounding LLMs as Efficient Decision-Making Agents via Offline Hierarchical Reinforcement Learning}

\begin{document}

\twocolumn[

% \icmltitle{Divide and Conquer: Building Efficient Decision-Making Agents via \\ Offline Hierarchical Reinforcement Learning}

\icmltitle{Divide and Conquer: Grounding LLMs as Efficient Decision-Making Agents \\ via Offline Hierarchical Reinforcement Learning}

\icmlsetsymbol{intern}{*}
\icmlsetsymbol{corr}{\Envelope}

\begin{icmlauthorlist}
\icmlauthor{Zican Hu}{nju,ailab,intern}
\icmlauthor{Wei Liu}{hkust}
\icmlauthor{Xiaoye Qu}{ailab}
\icmlauthor{Xiangyu Yue}{cuhk}
\icmlauthor{Chunlin Chen}{nju}
\icmlauthor{Zhi Wang}{nju,ailab,corr}
\icmlauthor{Yu Cheng}{cuhk,corr}
\end{icmlauthorlist}

\icmlaffiliation{nju}{Nanjing University}
\icmlaffiliation{ailab}{Shanghai AI Laboratory}
\icmlaffiliation{hkust}{The Hong Kong University of Science and Technology}

\icmlaffiliation{cuhk}{The Chinese University of Hong Kong}

\icmlcorrespondingauthor{Zhi Wang}{zhiwang@nju.edu.cn}
\icmlcorrespondingauthor{Yu Cheng}{chengyu@cse.cuhk.edu.hk}

\icmlkeywords{Machine Learning, ICML}
\vskip 0.3in]

\printAffiliationsAndNotice{}

\begin{abstract}
% While demonstrating impressive capabilities in language understanding, they often struggle with long-horizon decision-making tasks due to ineffective exploration and long-term credit assignment, especially in sparse reward scenarios.
% Inspired by the divide-and-conquer principle, we propose \textbf{GLIDER} (\textbf{G}rounding \textbf{L}anguage Models as Eff\textbf{I}cient \textbf{D}ecision-Making Agents via Offline Hi\textbf{E}rarchical \textbf{R}einforcement Learning), an innovative framework that introduces an easy-to-implement (or flexible or lightweight?) hierarchy to the LLM policy.
% We develop a scheme where the low-level controller is supervised with task-level plans that are learned and proposed automatically by the high-level policy.
% This design decomposes a complicated task instruction into a series of executable sub-tasks (akin to the chain-of-thought reasoning), significantly improving the ability of LLM policies to tackle complex long-horizon tasks with enhanced exploration.
% We first construct a base LLM policy via supervised fine-tuning on offline datasets, followed by training hierarchical actor-critic models with offline RL. 
% Moreover, we show that GLIDER can facilitate fast online adaptation to non-stationary environments owing to its strong transferability of low-level policies.
% Empirical studies demonstrate that GLIDER significantly improves efficiency and performance on ScienceWorld and ALFWorld benchmark tasks, attaining a sample efficiency of about ()x over existing methods.

While showing sophisticated reasoning abilities, large language models (LLMs) still struggle with long-horizon decision-making tasks due to deficient exploration and long-term credit assignment, especially in sparse-reward scenarios.
Inspired by the divide-and-conquer principle, we propose an innovative framework \textbf{GLIDER} (\textbf{G}rounding \textbf{L}anguage Models as Eff\textbf{I}cient \textbf{D}ecision-Making Agents via Offline Hi\textbf{E}rarchical \textbf{R}einforcement Learning) that introduces a parameter-efficient and generally applicable hierarchy to LLM policies.
We develop a scheme where the low-level controller is supervised with abstract, step-by-step plans that are learned and instructed by the high-level policy. 
This design decomposes complicated problems into a series of coherent chain-of-thought reasoning sub-tasks, providing flexible temporal abstraction to significantly enhance exploration and learning for long-horizon tasks.  
Furthermore, GLIDER facilitates fast online adaptation to non-stationary environments owing to the strong transferability of its task-agnostic low-level skills.
Experiments on ScienceWorld and ALFWorld benchmarks show that GLIDER achieves consistent performance gains, along with enhanced generalization capabilities.

% compared to various baselines.

\end{abstract}

\section{Introduction}
\label{intro}

\begin{figure}[!t]
\centering\includegraphics[width=0.85\columnwidth]{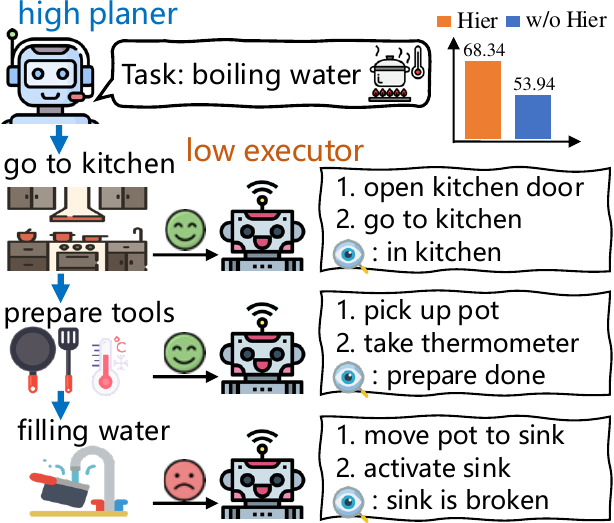}

\caption{
% GLIDER's hierarchical framework demonstrated through a "boiling water" example, showing improved performance over non-hierarchical approach. 
GLIDER's hierarchical framework, showing significant performance gain over non-hierarchical approaches. 
}
\vspace{-0.3cm}
\label{fig:idea}
\end{figure}

A longstanding goal of artificial general intelligence is to build agents capable of reasoning, decision-making, and communication~\citep{wooldridge1995intelligent,xu2024language}.
% With a wealth of semantic knowledge about the world, large language models (LLMs) have demonstrated remarkable potential in constructing intelligent agents across diverse domains, including reasoning ~\citep{wei2022chain,zhou2023leasttomost}, programming~\citep{yang2023intercode,luo2024wizardcoder}, task planning~\citep{lin2023swiftsage,valmeekam2023on}, embodied robotics~\citep{ahn2022can,shah2023lm}, and multi-agent environments~\citep{chen2024agentverse,ma2024large}. 
Recent attempts to exploit large language models (LLMs) as agents have shown commendable results in tackling interactive decision-making tasks~\citep{li2022pre,yao2023react,song2024trial}.
Prompt-based methods, like ReAct~\citep{yao2023react} and Reflexion~\citep{shinn2023reflexion}, recursively augment the prompt to a frozen LLM with verbal feedback.
They are prone to exceed the input length limit of in-context learning, especially for long-horizon tasks.
% They can easily exceed the input length limit of in-context learning, especially for complex tasks with long-horizon interactions.
% Scaling with fine-tuning techniques can further unlock the potential of LLMs for downstream applications. The standard approach is behavior cloning that employs a supervised teacher-forcing manner to fine-tune LLM policies on expert demonstrations~\citep{zeng2023agenttuning,chen2023fireact,lin2023swiftsage}. 
Scaling with supervised fine-tuning techniques can further unlock the potential of LLMs for downstream applications~\citep{zeng2023agenttuning,chen2023fireact,lin2023swiftsage}.
However, their performance is highly dependent on expensive expert demonstrations and can be limited due to deficient exploration of target environments.

% Intelligent agents must excel not only at mimicking successful trials but also at autonomous behavior adaptation through trial-and-error interactions
Intelligent agents must excel at both imitating demonstrations and adapting behaviors through trial-and-error~\citep{silver2021reward,rafailov2023direct}.
Modern approaches adopt reinforcement learning (RL) algorithms to steer LLMs toward user-specified tasks~\citep{ouyang2022training}, such as offline Q-learning~\citep{snell2023offline}, PPO~\citep{zhai2024fine,szot2024large}, or DPO~\citep{song2024trial}.
% This paradigm emerges as the key component in achieving SOTA performance for advanced LLMs like OpenAI o1 and DeepSeek-R1~\citep{guo2025deepseek}.
This paradigm enables SOTA performance in advanced LLMs like OpenAI o1 and DeepSeek-R1~\citep{guo2025deepseek}.
However, RL intrinsically requires tedious and vast environment interactions, leading to brittle performance and poor sample efficiency~\citep{burda2019exploration,mahankali2024random}.
Building efficient LLM agents with open-ended textual commands poses several challenges, such as tackling huge action spaces, executing long-horizon planning, and learning from sparse-reward feedback~\citep{rocamonde2024vision,dwaracherla2024efficient}.
Existing works still struggle with complex tasks that demand a broad spectrum of vital capabilities, including long-term credit assignment, understanding the real physical world, and sophisticated exploration with structured reasoning~\citep{qiao2024agent,zhou2024archer}.

% In contrast, humans naturally tackle complex problems through hierarchical decomposition. This principle is evident across scales: from how corporations divide into specialized departments to how biological systems organize from cells to tissues to organs. 
Humans naturally tackle complex problems through hierarchical decomposition~\citep{sutton1999between}.
This divide-and-conquer principle is evident across scales in natural systems, from how corporations divide into specialized departments to how biological systems organize cells to form tissues and organs.
% Such hierarchical approaches consistently demonstrate remarkable efficiency in managing complex tasks, suggesting a promising direction for artificial decision-making systems.
The hierarchy design plays a crucial role in advancing frontier research on language agents~\citep{li2024optimus,zhou2024archer} and embodied intelligence~\citep{ahn2022can,black2024pi}, showcasing remarkable efficiency for solving intricate tasks in a more human-like manner.

% frontier
% Hierarchical methods play a crucial role in advancing frontier research on language agents, as they enable more structured and efficient problem-solving by breaking down complex tasks into manageable subtasks, thereby enhancing both performance and interoperability

% Hierarchical methods play a pivotal role in advancing the frontier of language agents, enabling them to effectively navigate complex decision-making tasks and interact with their environment in a more human-like manner."
% The incorporation of hierarchical methods is crucial in pushing the boundaries of language agent research, as it allows these models to better generalize across diverse scenarios and tasks, thereby enhancing their overall capabilities."
% By leveraging hierarchical methods, researchers can significantly improve the performance and adaptability of language agents, ultimately paving the way for more sophisticated and human-like interactions in various applications.".

Inspired by this, we propose an innovative framework \textbf{GLIDER} (\textbf{G}rounding \textbf{L}anguage Models as Eff\textbf{I}cient \textbf{D}ecision-Making Agents via Offline Hi\textbf{E}rarchical \textbf{R}L) that introduces a parameter-efficient and generally applicable hierarchy to train competent LLM policies for complex interactive tasks. 
Our scheme contains two LLM policies, where the low-level controller is supervised to achieve abstract, step-by-step plans learned and proposed by the high-level instructor. 
By harnessing the strong reasoning and planning capabilities of LLMs, we can decompose complicated problems into a series of coherent chain-of-thought (CoT) reasoning sub-tasks and perform efficient exploration in a semantically structured space. 
To enhance learning stability, we first build a base agent via behavior cloning, followed by reinforcement fine-tuning of the hierarchical token-level actors and sentence-level critics.
This two-level agent is trained in offline mode using hierarchical datasets to achieve prominent sample efficiency, and also can be seamlessly deployed for offline-to-online fine-tuning scenarios.
% \textcolor{red}{introduce experiments, significant performance, surpass baseline xxx points...}

In summary, our main contributions are as follows:
\begin{itemize}[itemsep=0.25em, leftmargin=1.25em]\vspace{-0.5em}
    
    % \item We propose an offline hierarchical framework with parameter efficiency and broad applicability, empowering LLM agents to tackle complex decision-making tasks via sophisticated exploration and structured reasoning.

    \item Inspired by the divide-and-conquer principle, we propose an offline hierarchical framework GLIDER 
    %with superior parameter efficiency and broad applicability
    , empowering LLM agents to tackle complex decision-making tasks via sophisticated exploration and structured reasoning.

    \item Our method enables fast offline-to-online adaptation to non-stationary environments by developing highly generalizable skills through hierarchical LLM agents.

    % \item Comprehensive experiments on ScienceWorld and ALFWorld benchmarks show that our method consistently achieves superior performance and sample efficiency, as well as improved generalization capacity.

    \item Comprehensive studies on ScienceWorld and ALFWorld benchmarks show that our method consistently improves performance and generalization capacity, surpassing a range of baselines by a significant margin.
    
\end{itemize}

\begin{figure*}[tb]
\centering\includegraphics[width=0.98\textwidth]{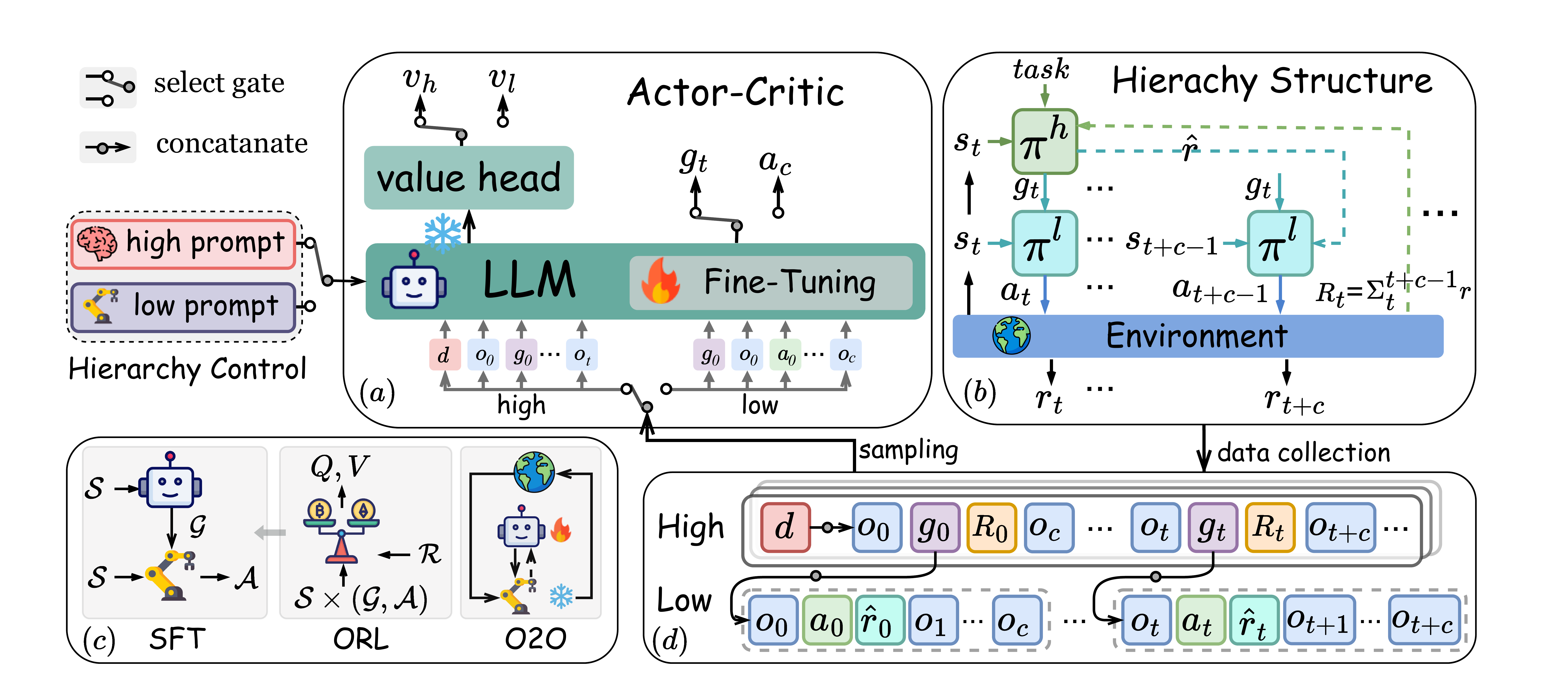}
\caption{Overview of the GLIDER framework. 
% (a) Actor-Critic architecture with prompt-controlled high and low-level training on sampled trajectories from offline dataset. 
(a) Hierarchical Actor-Critic architecture with prompt-controlled high- and low-level training on sampled trajectories from offline datasets. 
% (b) Hierarchical policy for data collection and evaluation interaction with the environment, where $\pi^h$ generates the high-level goals and  $\pi^l$ executes low-level actions. 
(b) Hierarchical policy structure where the high-level $\pi^h$ generates sub-task $g$ only when the low-level $\pi^l$ executes primitive actions for $c$ steps. 
The high-level policy provides the low-level with an intrinsic reward $\hat{r}$ that indicates the sub-task completion, and collects environment rewards across $c$ timesteps as its one-time reward as $R_t\!=\!\Sigma r_{t:t+c-1}$.
% (c) The training pipeline comprises SFT, ORL (offline RL), O2O (offline-to-online RL) stages. 
(c) The training pipeline comprises SFT, ORL (offline RL), and O2O (offline-to-online RL) stages. 
% (d) Structured hierarchical trajectories composed of high-level tuples $(d,o,g,R)$ for task description, observation, subtask, and high-level reward, where $R=\sum{r}$, and low-level tuples $(g,o,a,\hat{r})$ for subtask, observation, action, and low-level reward.
(d) Structured hierarchical trajectories composed of high-level transitions $(d;o_t,g_t,R_t,o_{t+c})$ and low-level transitions $(g; o_t,a_t,\hat{r}_t,o_{t+1})$.
}
\label{framwork}
\end{figure*}

\section{Related Work}
\label{related}

\textbf{LLMs as Decision-Making Agents.}
% By harnessing the reasoning and in-context learning ability of LLMs, building decision-making agents with LLMs is attracting increasing interest over various domains~\citep{xi2023rise}, such as program coding~\citep{yang2023intercode,luo2024wizardcoder}, web surfing~\citep{nakano2021webgpt,deng2023mind2web}, multi-agent games~\citep{ma2024large,xu2024language}, and real-world applications~\citep{ahn2022can,black2024pi}.
% Building decision-making agents with LLMs is attracting increasing interest across various domains~\citep{xi2023rise,black2024pi}.
With a wealth of semantic knowledge about the world, LLMs have shown remarkable potential in building competent agents across diverse domains~\citep{xi2023rise}, including reasoning ~\citep{wei2022chain,zhou2023leasttomost,luo2024wizardcoder}, robotics~\citep{ahn2022can,shah2023lm} and multi-agent~\citep{chen2024agentverse,ma2024large}.
% programming~\citep{yang2023intercode, luo2024wizardcoder}, task planning~\citep{lin2023swiftsage,valmeekam2023on} and multi-agent environments~\citep{chen2024agentverse,ma2024large}. 
% Early works freeze pretrained LLMs and make decisions under a prompt-based framework.
% CoT methods~\citep{wei2022chain,yao2023tree,wang2023plan} prompt LLMs with a series of intermediate reasoning steps, asking them to think step-by-step.
Early studies use a prompt-based framework, such as classical CoT methods~\citep{wei2022chain,yao2023tree,wang2023plan} that prompt LLMs with intermediate reasoning steps.
% ReAct~\citep{yao2023react} generates interleaved reasoning traces and action plans, allowing for structured synergy between reasoning and acting. Follow-up research continues to enhance performance by incorporating self-reflective verbal feedback~\citep{shinn2023reflexion} or using strategic reasoning~\citep{gandhi2023strategic}.
Follow-up research employs synergy between reasoning and acting~\citep{yao2023react}, incorporates self-reflective verbal feedback~\citep{shinn2023reflexion}, and uses strategic reasoning~\citep{gandhi2023strategic}.
They use recursive feedback traces to augment prompts, which helps address long sequence and complex, long-horizon task challenges.
% They recursively augment the prompt with feedback traces, which can easily exceed the input length limit of in-context learning, especially for complex, long-horizon tasks.

% Scaling with fine-tuning techniques can unlock the potential for downstream tasks. The standard option is supervised behavior cloning that fine-tunes LLM policies on expert demonstrations~\citep{hu2022lora}.
% Existing studies continually increase the capacity of LLM agents by fine-tuning them on oracle action trajectories~\citep{lin2023swiftsage}, on a lightweight instruction-tuning dataset containing high-quality interaction trajectories~\citep{zeng2023agenttuning}, and on agent trajectories generated from multiple tasks and prompting methods~\citep{chen2023fireact}.
Scaling with supervised fine-tuning techniques can unlock the potential for downstream tasks~\citep{hu2022lora}, such as fine-tuning LLM agents on oracle action trajectories~\citep{lin2023swiftsage}, on a lightweight instruction-tuning dataset containing high-quality interaction trajectories~\citep{zeng2023agenttuning}, and on agent trajectories generated from multiple tasks and prompting methods~\citep{chen2023fireact}.
Due to inherent limitations, the performance relies heavily on expensive high-quality data and could easily be restricted by deficient exploration of target environments.

RL offers a natural paradigm to unleash the LLM agents' decision-making capabilities~\citep{ouyang2022training}.
\citet{snell2023offline} guides language generation towards maximizing user-specified utility functions using implicit language Q-learning.
% ETO~\citep{song2024trial} interacts with the environment to obtain contrastive trajectory pairs and utilizes these preference pairs to update the LLM policy using DPO~\citep{rafailov2023direct}.
ETO~\citep{song2024trial} collects contrastive trajectory pairs from interactions to update the LLM policy using DPO~\citep{rafailov2023direct}.
Other works employ a similar pipeline where an LLM policy interacts with the environment to receive goal-directed task rewards, which are then used to fine-tune the policy with classical algorithms like PPO~\citep{zhai2024fine,szot2024large,tan2024true}.
Building LLM agents with open-ended textual commands can involve a huge action space, and long-horizon planning or sparse-reward scenarios~\citep{mahankali2024random,zhou2024archer}.
This motivates us to introduce a hierarchy to ground LLMs as efficient decision-making agents.

\textbf{Hierarchical RL} emerges as a powerful framework for managing complexity in decision-making tasks.
Classical approaches of Options~\citep{sutton1999between,bacon2017option} and MAX-Q~\citep{dietterich2000hierarchical} formalize temporal abstractions in RL. 
Recent advances have significantly expanded these foundations, such as HiRO~\citep{nachum2018data} with data-efficient off-policy training for hierarchical policies, HAC~\citep{levy2019learning} with parallel training of 3-level hierarchies, and HiPPO~\citep{hippo2020} with efficient hierarchical policy gradient approximation for robust skill training.
In general, a persistent challenge is the dependence on domain expertise to specify meaningful hierarchies.
This limitation also motivates us to harness the strong semantic understanding and reasoning abilities of LLMs for natural task decomposition with an autonomous hierarchy.

\textbf{Offline-to-Online RL.}
Offline RL~\citep{levine2020offline} harnesses offline data without environment interactions, yielding effective algorithms such as BCQ~\citep{fujimoto2019off}, CQL~\citep{kumar2020conservative}, and Fisher-BRC~\citep{kostrikov2021offline}.
Also, it remains beneficial to fine-tune the pretrained offline policy with further online interactions, that is, offline-to-online RL~\citep{nair2020awac}.
However, such a benefit is often diminished due to remarkable distribution shifts between pretraining and deployment, leading to accumulated bootstrap errors~\citep{lee2022offline,wang2023train}.
Moreover, existing works often require extensive retraining for new tasks~\citep{yu2023actor}.
% In the paper, we show that our method achieves efficient offline-to-online adaptation to non-stationary environments. In the offline stage, we decompose the LLM policy into low-level skills with physical semantic meaning and a high-level instructor with strong reasoning abilities. With these highly generalizable skills that are robust to distribution shift, we can efficiently fine-tune the high-level planner for fast online adaptation.
In contrast, we show that our method achieves fast online adaptation to non-stationary environments via training highly generalizable low-level skills that are robust to distribution shifts.

\iffalse 
\section{Preliminaries: LLM as Policies}
\label{pre}

We formulate RL as a standard Markov decision process (MDP) with a tuple $\langle \mathcal{S}, \mathcal{A}, \mathcal{T}, \mathcal{R}, \gamma \rangle$, where $\mathcal{S}/\mathcal{A}$ is the state/action space, $\mathcal{T}/\mathcal{R}$ is the state transition/reward function, and $\gamma\in(0, 1]$ is the discount factor.
We formulate an LLM-based policy, $\pi_{\theta}:\mathcal{S}\times\mathcal{A}\to [0,1]$, as a probability distribution that maps states to actions, where both $\mathcal{S}$ and $\mathcal{A}$ are drawn from text spaces constrained by user-specified tasks, and policy parameters $\theta$ are initialized from a pretrained LLM.
The objective is to optimize the policy to receive a maximal expected return as $J(\pi)=\mathbb{E}_{\pi}[\sum_t\gamma^tr_t]$.

% Offline RL, also known as batch RL, is a type of RL where an agent learns to make decisions by analyzing a fixed dataset of previously collected experiences, rather than interacting with an environment in real-time. In other words, the agent learns from a batch of offline data rather than actively exploring and collecting new data online.
\fi

% \section{Methodology}
\section{Method}
\label{sec:method}

In this section, we present GLIDER, which introduces a parameter-efficient and generally applicable hierarchy to ground LLMs as efficient decision-making agents for tackling complex interactive tasks.
Figure~\ref{framwork} illustrates the framework, containing a three-stage pipeline of base agent construction via supervised fine-tuning, policy refinement via offline RL, and seamless adaptation to online deployment.
The algorithm pseudocode is given in Appendix~A, and detailed implementations are presented as follows.

\subsection{Problem Setup of the LLM Agent}
We formulate the agent task as a standard Markov decision process (MDP) with a tuple $\langle \mathcal{S}, \mathcal{A}, \mathcal{T}, \mathcal{R}, \gamma \rangle$, where $\mathcal{S}/\mathcal{A}$ is the state/action space, $\mathcal{T}/\mathcal{R}$ is the state transition/reward function, and $\gamma\in(0, 1]$ is the discount factor.
We introduce an LLM-based policy, $\pi_{\theta}:\mathcal{S}\times\mathcal{A}\to [0,1]$, a probability distribution that maps states to actions, where both $\mathcal{S}$ and $\mathcal{A}$ are drawn from text spaces constrained by user-specified tasks, and policy parameters $\theta$ are initialized from a pretrained LLM.
The objective is to optimize the policy to receive a maximal expected return as $J(\pi)=\mathbb{E}_{\pi}[\sum_t\gamma^tr_t]$.

% \subsection{Hierarchical LLMs-based Agent Architecture}
\subsection{Hierarchical Architecture of the LLM Agent}
We extend the LLM agent setup to a hierarchical two-layer structure, with a high-level policy $\pi_{\theta}^h$ and a low-level controller $\pi_{\theta}^l$.
The high-level policy operates at a coarser layer of task planning and sets sub-task goals for the low-level to accomplish.
At each timestep $t$, the environment provides an observation $o_t$.
The high-level planner $\pi_{\theta}^h$ receives the observation $o_t$ together with a textual task description $d$, and produces a high-level goal (or sub-task) as $g_t\sim \pi_{\theta}^h(\cdot\mid d, o_t)$ when $t\equiv 0~(\text{mod}~c)$.
This provides temporal abstraction, since high-level decisions are made only every $c$ steps.
\footnote{$c$ could differ across sub-tasks, as harder sub-tasks naturally require more primitive actions to accomplish.}
For the next $c$ timesteps, the low-level controller receives the environment observation $o_t$ and goal $g_t$, and produces a low-level primitive action as $a_t \sim \pi^l_{\theta}(\cdot \mid g_t, o_t)$.
The action $a_t$ is applied to the environment, which yields a reward $r_t$ and transitions to a new observation $o_{t+1}$.
The high-level policy collects environment rewards through these $c$ timesteps, storing the high-level transition $(d;~o_t, g_t, \Sigma r_{t:t+c-1}, o_{t+c})$ for offline training.
Correspondingly, the high-level dataset for offline learning stages is constructed as 
\begin{eqnarray}\label{data_high}
    \mathcal{D}^h=\Sigma_N \left[~d;\right. &&\!\!\!\!\!\!\!\!\!\!\! \left(o_0, g_0, \Sigma r_{0:c-1}, o_{c}\right),..., \nonumber\\
    &&\!\!\!\!\!\!\!\!\!\!\! \left.\left(o_t, g_t, \Sigma r_{t:t+c-1}, o_{t+c}\right),...~\right],
\end{eqnarray}
which captures strategic task planning over $N$ trajectories.

The high-level policy provides the low-level with an intrinsic reward $\hat{r}$ that indicates the sub-task completion ($1$ when the sub-task is completed and $0$ otherwise).\footnote{In a boiling water task, the high-level policy decomposes it into atomic subtasks (e.g., navigation, tools preparation). For navigation subtask, the low-level policy receives a reward of 1 upon reaching kitchen (verified through observation), 0 otherwise.}
The sub-task completion can be easily accessible from the environment observation $o_t$, without requiring any manual design or domain knowledge.

Each generated goal $g_t$ corresponds to a sequence of $c$ atomic transitions in the low-level dataset as 
\begin{eqnarray}\label{data_low}
\mathcal{D}^l=\Sigma_N\Sigma_t\left[~g_t;\right. &&\!\!\!\!\!\!\!\!\!\!\! \left(o_{t},a_{t},\hat{r}_{t},o_{t+1}\right),..., \nonumber \\
&&\!\!\!\!\!\!\!\!\!\!\! \left.\left(o_{t+c-1},a_{t+c-1},\hat{r}_{t+c-1},o_{t+c}\right)~\right],
\end{eqnarray}

\textbf{We design a parameter-efficient hierarchical model architecture.}
As shown in Figure~\ref{framwork}-(a), our model offers superior parameter efficiency from two perspectives.
First, the actor and critic share the same frozen LLM backbone, each introducing a minimal number of parameters for efficient fine-tuning at a lightweight computing cost.
The actor is formed by augmenting the backbone with LoRA~\citep{hu2022lora}, which adds a trainable low-rank bypass to each transformer block.
The critic is constructed by adding additional MLP layers to the last transformer block of the backbone.
Second, the high- and low-level policies share the same actor-critic models, differing only in a \textit{hierarchy prompt} that specifies the level of current inputs.
This design benefits from harnessing the powerful capability of LLMs to perform in-context learning, i.e., tackling a series of complex tasks by feeding short prompts to a single foundation model~\citep{brown2020language}.
In contrast, traditional hierarchical methods usually train independent models at each level, resulting in a multiplication of model parameters~\citep{nachum2018data,levy2019learning,hippo2020}.

\textbf{Our hierarchy setup achieves broad applicability.}
The hierarchical structure provides temporal abstraction with efficient exploration since high-level decisions are made only when the low-level controller executes for several steps.
The high-level policy unlocks the chain-of-thought reasoning of LLMs to decompose a complicated task into a series of coherent sub-task plans, while the low-level model translates abstract plans into precise, executable atomic actions. 
Our setup achieves generality by training the low-level policy to accomplish sub-task goals learned and instructed by the high-level planner.
The high-level planner is guided by environment-provided rewards, while the low-level policy is instructed by the sub-task completion signal derived from environment observations.
The whole learning process eliminates the necessity for any manual or task-specific design, making it broadly applicable.

% \subsection{Multi-stage Training Pipeline}
\subsection{Base Agent Construction via Behavior Cloning}\label{bc}
% \paragraph{Expert Imitation through Behavior Cloning.}
% Directly deploying LLMs as executors may generate invalid or inconsistent actions for downstream tasks due to the discrepancy between natural language space and environment-specific action space. 
Directly deploying LLMs as decision-making agents in downstream tasks may generate hallucinatory or inconsistent actions and perform brainless trial-and-error due to the semantic discrepancy between natural language (trained with next-token prediction) and user-specified environments.
% To bridge this gap, we initialize our approach with behavior cloning through supervised fine-tuning, leveraging demonstration trajectories to align the policy with valid action sequences.
To improve learning stability and sample efficiency, we first construct a base LLM agent through supervised fine-tuning of the hierarchical actors using pre-collected demonstration trajectories.
This behavior cloning process aligns the initial policy with valid action sequences and serves as a solid starting point for building a powerful agent.

Specifically, we imitate the behavior patterns, i.e., the state-to-action mapping function, within both levels of datasets in Eqs. (\ref{data_high}) and (\ref{data_low}).
The objective function of behavior cloning via supervised fine-tuning is to maximize the log-likelihood of the observed data.
Pre-trained LLMs tend to generate lengthy, verbose sentences that could be redundant and challenging to understand in the user-specified environment.
Hence, we incorporate a length regularization term to encourage the LLM policy to generate concise task plans and atomic actions for effective interaction with the environment.
The final loss function is formulated as 
\begin{equation}\label{bc_loss}
\begin{aligned}
    \mathcal{L}_{\text{SFT}}(\theta) = & - \mathbb{E}_{(d,o;g) \sim \mathcal{D}^h} \left[\log \pi_{\theta}^h(g|d,o) \right] + \lambda\cdot n_h  \\
    & -  \mathbb{E}_{(g,o;a) \sim \mathcal{D}^l} \left[\log \pi_{\theta}^l(a|g,o)\right] + \lambda\cdot n_l,
\end{aligned}
\end{equation}
where $\lambda$ is the length regularization ratio, and $n_h/n_l$ is the output length of the high-/low-level policy.

% To prevent the policy from excessively increasing the sequence length to achieve higher action probabilities, we incorporate a length regularization term into the objective function as:
% \begin{equation}\label{regular}
%    \log \pi(a|s) =\sum_{i=1}^{l} \log \pi(w^i \mid s,w^{1:i-1}) - \lambda \cdot l
% \end{equation}
% where $\lambda$ is the length regularization hyperparameter.

% The objective function for behavior cloning is formulated as follow:
% \begin{equation}\label{bc_loss}
%     \mathcal{L}_{\theta} = -\mathbb{E}_{(s,a) \sim \mathcal{D}} \log \pi_{\theta}(a|s)
% \end{equation}

% By employing behavior cloning (BC) for initial policy learning, we establish a robust foundation for generating valid actions. 
By employing behavior cloning for a base LLM agent construction, we establish a robust foundation for generating valid actions with significantly improved sample efficiency.
% However, it is crucial to recognize that this initial BC-derived agent, while capable of producing valid actions, remains fundamentally limited in addressing complex long-horizon decision-making challenges. 
However, the base agent is easily limited by the data quality of demonstration trajectories and lacks the ability to explore the environment.
It could easily result in sub-optimal policies, especially when tackling complex long-horizon decision-making challenges.
% Consequently, we subsequently employ reinforcement learning (RL) techniques to refine and enhance the policy's capabilities, enabling the agent to navigate more sophisticated, extended decision-making scenarios that extend beyond the initial behavioral cloning framework.

% To train a more powerful agent, it is important for the model to also explore failure trajectories. To achieve this, a viable approach is reinforcement learning

\subsection{Offline Hierarchical Policy Refinement}
To unlock the capacity of LLM agents in long-horizon decision-making, we continue to train hierarchical actor-critic models in an offline mode using the reward-annotated datasets $\mathcal{D}^h$ and $\mathcal{D}^l$.
The actor outputs a sequence of tokens autoregressively for fine-grained action generation and control at the \textit{token level}, while the critic aims to evaluate the output policy at the \textit{sentence level}.
In the following, we use $s$ and $u$ to denote the state and action uniformly, i.e., $s=(d,o), u=g$ for the high-level policy and $s=(g,o), u=a$ for the low-level.

\textbf{Sentence-Level Critic.}
% The critic component consists of a Q-function $Q_{\phi}(s,a)$ and a value function  $V_{\psi}(s)$ that are trained using temporal difference learning. 
Following the practice in advanced offline RL algorithms~\citep{snell2023offline,zhou2024archer}, the critic component consists of a Q-function $Q_{\phi}(s,u)$ and a value function $V_{\psi}(s)$ that are optimized using temporal difference learning.
% The Q-function is trained to minimize the mean squared Bellman error:
% \begin{equation}\label{q_value}
%     \mathcal{J}_Q(\phi) = \mathbb{E}_{(s,a,r,s')\sim \mathcal{B}}\left[\left(Q_{\phi}(s,a) - r - \gamma V_{\bar{\psi}}(s')\right)^2\right]
% \end{equation}
% where $\phi$ represents the Q-function parameters, $\mathcal{B}$ is the offline dataset, and $V_{\bar{\psi}}$ is the target value network with frozen parameters $\bar{\psi}$.
The Q-function is trained to minimize the Bellman bootstrapping error as 

\begin{equation}\label{q_value}
     \mathcal{L}_Q(\phi) \!=\! \mathbb{E}_{(s,u,r,s')\sim D_r}\!\!\left[\left(r + \gamma V_{\bar{\psi}}(s')-Q_{\phi}(s,u) \right)^2\right]\!,\!
\end{equation}
where the Bellman target is computed from a delayed copy of the value model $V_{\bar{\psi}}$.
% The value function is trained using an asymmetric loss function to maintain a conservative estimate:
% \begin{equation}\label{value}
%     \mathcal{J}_V(\psi) = \mathbb{E}_{s\sim \mathcal{B}}\left[\mathbb{E}_{a \sim \pi_\theta(\cdot \mid s)}\left[L_2^\tau \left( V_{\psi}(s) - Q_{\bar{\phi}}(s,a) \right)\right] \right]
% \end{equation}
% where $L_2^\tau(u)|\tau - \mathbbm{1} (u<0)| u^{2}$ is an asymmetric loss function with parameter $\tau \in [0.5, 1)$. This loss function penalizes overestimation more heavily than underestimation: when $V_{\psi}(s) > Q_{\bar{\phi}}(s,a)$ (overestimation), the loss is weighted by $\tau$; when $V_{\psi}(s) < Q_{\bar{\phi}}(s,a)$ (underestimation), the loss is weighted by $(1-\tau)$.
The value function is trained using an asymmetric loss function to maintain a conservative value estimation as 
\begin{equation}\label{value}
    \mathcal{L}_V(\psi) \!=\! \mathbb{E}_{s\sim D_r}\!\!\left[\mathbb{E}_{u \sim \pi_\theta(\cdot \mid s)}\!\left[L_2^\tau \!\left(Q_{\bar{\phi}}(s,u) \!-\! V_{\psi}(s) \right)\right] \right]\!,\!
\end{equation}
where $L_2^\tau(x)=|\tau - \mathbbm{1} (x<0)| x^{2}$ is an asymmetric loss function with a expectile parameter $\tau \in [0.5, 1)$, introduced in implicit Q-learning~\citep{kostrikov2022offline}. 
% This loss function penalizes overestimation (when $V_{\psi}(s) > Q_{\bar{\phi}}(s,u)$, weighted by $\tau$) more heavily than underestimation (when $V_{\psi}(s) < Q_{\bar{\phi}}(s,u)$, weighted by $1-\tau$).
This loss function assigns more importance to $Q\!>\!V$ predictions (weighted by $\tau$) while reducing the influence of $Q\!<\!V$ predictions (weighted by $1\!-\!\tau$).
The asymmetric design helps prevent the learned value function from being overly optimistic, as overestimation could easily lead to poor policy updates in offline RL settings due to distribution shift. 
The delayed target networks $\bar{\psi}$ and $\bar{\phi}$ are periodically updated using Polyak averaging~\citep{haarnoja2018soft} to improve training stability.

\textbf{Token-Level Actor.}
% Policy optimization aims to maximize expected discounted return, $J(\pi) = \mathbb{E}_{\tau \sim \pi}[\sum_{t=0}^T \gamma^t r_t]$, where $\gamma \in [0,1]$ is the discount factor, $\tau$ is a trajectory, and $r_t$ is the reward at time step $t$. The policy optimization objective can be formulated as:
% \begin{equation}\label{policy}
% \begin{aligned}
%     \mathcal{J}_\pi(\theta) 
%     & = -\mathbb{E}_{(s,a) \sim \mathcal{B}} \left[\exp(\frac{1}{\lambda} A(s,a))\log \pi_\theta (a \mid s) \right]
% \end{aligned}
% \end{equation}
The LLM-based actor outputs a sequence of tokens $w_{1:n}$ autoregressively, where $n$ is the output sentence length.
Each token is selected according to the token probability distribution generated by token actor as 
\begin{equation}\label{policy_prob}
    \pi_{\theta}(u\mid s)=\pi_{\theta}(w_{1:n}\mid s)=\prod_{i=1}^{n}  \pi_{\theta}(w_i \mid s, w_{1:i-1}).
\end{equation}
The actor is trained to maximize the expected return of the policy, i.e., the estimated Q-function, which is also equivalent to maximizing the advantage function.
Following the practice in AWAC~\citep{nair2020awac}, we formulate the policy optimization of the token actor as a weighted maximum likelihood estimation problem. 
The resulting loss function is derived as 
\begin{eqnarray}\label{awac}
     \mathcal{L}_\pi(\theta)  &&\!\!\!\!\!\!\!\!\!\! = -\mathbb{E}_{(s,u) \sim D_r} \left[\exp\left(\frac{1}{\lambda} A(s, u)\right)\cdot \log \pi_\theta (u \mid s) \right] \nonumber\\
    &&\!\!\!\!\!\!\!\!\!\! = -\mathbb{E}_{(s,u) \sim D_r} \left[\exp \left(\frac{1}{\lambda} \left( Q_{\phi}(s,u)-V_{\psi}(s) \right) \right) \right. \\
    &&~~~~~~~~~~~~~~~~~~~~ \left. \cdot \sum_{i=1}^n\log \pi_\theta (w_i \mid s, w_{1:i-1}) \right]. \nonumber 
\end{eqnarray}

This ``supervised" formulation implicitly enforces a constraint to mitigate distribution shift and avoids overly conservative updates with advantage weighting, thus facilitating efficient hierarchical policy learning from offline data.
Further, by eliminating over-conservatism and explicit modeling of the behavior policy, it is well suited to perform fast adaptation to new tasks in online deployment, as studied in Sec.~\ref{o2o}.

\subsection{Offline-to-Online Adaptation}\label{o2o}
% During Offline-to-Online(O2O) deployment, GLIDER can efficiently adapt to changes in the environment or task by leveraging the flexibility of its hierarchical structure. Notably, in online operation, the low-level agent remains fixed because the underlying skills for new tasks are often similar. Therefore, we only need to adjust the high-level planner.
% This targeted fine-tuning allows for rapid modifications to task-level plans based on new inputs without necessitating changes to the low-level controller. As a result, GLIDER can quickly adapt to new tasks while maintaining strong performance, making it well-suited for dynamic, non-stationary environments. This efficient approach minimizes the computational overhead typically associated with retraining the entire model.
In the offline stage, we decompose the LLM agent into a series of low-level sub-tasks (or skills) and a high-level policy with strong reasoning abilities.
With this flexible hierarchical structure, GLIDER can be efficiently adapted to new environments with further online interactions in offline-to-online scenarios.
The low-level skills are pre-trained using intrinsic reward functions rather than task-specific ones, allowing for high generalization capacity across tasks and good robustness to the distribution shift between offline pre-training and online deployment.
% Naturally, we fix the task-agnostic low-level skills and only fine-tune the high-level policy with online interactions with the new environment.
Naturally, we freeze the task-agnostic low-level skills that interact with the new environment, and only fine-tune the high-level policy with the environment-provided reward signals.

Formally, at each timestep $t$, the high-level policy receives the environment observation $o_t$ together with the task description $d$, and selects a low-level skill as $g_t\sim \pi_{\theta}^h(\cdot | d, o_t)$.
Then, the fixed skill $g_t$ interacts with the environment for $c$ primitive actions, resulting in a $c$-step trajectory as $[g_t;~(o_t,a_t,r_t,o_{t+1}),...,(o_{t+c-1},a_{t+c-1},r_{t+c-1},o_{t+c})]$.
We construct the transition sample for the high-level policy as $(d;~o_t, g_t, \Sigma r_{t:t+c-1}, o_{t+c})$.
Finally, we collect these transition samples to fine-tune the high-level critic with Eqs.~(\ref{q_value})-(\ref{value}) and the actor with Eq.~(\ref{awac}).
By harnessing the temporal abstraction knowledge embodied in the pretrained low-level skills, GLIDER can quickly adapt to non-stationary environments with significantly improved exploration efficiency.

% In the offline stage, we decompose the LLM policy into low-level skills with physical semantic meaning and a high-level instructor with strong reasoning abilities.
% With these highly generalizable skills that are robust to distribution shift, we can efficiently fine-tune the high-level planner for fast online adaptation.

% The full GLIDER algorithm for multi-stage training pipline is summarized in Algorithm\ref{algo}.

\section{Experiments}
We evaluate GLIDER in offline settings from Sec.~\ref{sec:performance} to Sec.~\ref{sec:data_rate}, and then test its adaptability in an online fine-tuning manner in Sec.~\ref{sec:generalization}.
Through comprehensive experiments, we aim to answer the following research questions:
\begin{itemize}[itemsep=0.25em, leftmargin=1.25em]\vspace{-0.5em}
    % \item How effective and robust is GLIDER across diverse settings? We examine its performance against prompt-based methods and fine-tuning baselines, assess consistency across different backbone models, and evaluate adaptability in both sparse and dense reward environments. (See Sec.~\ref{sec:performance}).
    \item How effective and robust is GLIDER across diverse settings? We examine its performance against prompt-based and fine-tuning baselines, assess consistency across different backbones, and evaluate agent capacity in both sparse and dense reward environments. (See Sec.~\ref{sec:performance}).

    \item What is the contribution of each component to GLIDER's performance? Through systematic ablation studies, we analyze the impact of the hierarchical structure, training stages (SFT and ORL), and variations in model architecture and scale. (See Sec.~\ref{sec:ablation}).

    \item How well does GLIDER generalize to out-of-domain tasks through online fine-tuning? We evaluate the model's adaptation capabilities on previously unseen task distributions. (See Sec.~\ref{sec:generalization}).

    % \item How different ratios of expert demonstrations to medium-quality data affect model performance? We compare the different mixture strategy for training data composition. ( (See Sec.~\ref{sec:data_rate}).
    \item How do varying ratios of expert demonstrations to medium-quality data affect model performance? We evaluate different mixture strategies for the composition of training data. (See Sec.~\ref{sec:data_rate}).
\end{itemize}

\label{exp}
\subsection{Experimental Settings}
\paragraph{Benchmarks and Offline Dataset.} 
% We evaluate GLIDER on two language-base interactive decision-making tasks:
We evaluate GLIDER on two popular language-based interactive decision-making tasks:
% 1) \texttt{ScienceWorld} \citep{wang2022scienceworld} is a language-based environment for elementary science experiments, featuring 30 tasks across 10 categories. 
1) \texttt{ScienceWorld}~\citep{wang2022scienceworld} is a textual environment for elementary science experiments, featuring 30 tasks. 
% Agents must demonstrate scientific understanding through interactive experimentation, with progress measured by a dense reward (0 to 1) at each step.
% 2) \texttt{ALFWorld}~\citep{shridhar2021alfworld} simulates household environments that require navigation and object manipulation.
 % simulates household environments that require navigation and object manipulation in a sparse, binary reward setting.
2) \texttt{ALFWorld}~\citep{shridhar2021alfworld} contains 6 types of household manipulation tasks, requiring agents to navigate and interact with objects following language instructions in a binary reward setting.
% Task success is measured through a binary reward system (0 or 1).
% The reward is 1 only upon successful task completion, and 0 otherwise.
% Beyond standard evaluation on seen tasks, it includes unseen scenarios to assess generalization ability. 
% For details of the expert and medium trajectory collection process, and hierarchical dataset organization can be found in appendix.

For offline training data, we construct a dataset that combines expert demonstrations (optimal trajectories provided by benchmarks) and medium-quality trajectories with a mixture ratio of $1:2$. The medium-quality trajectories are collected through two strategies: in-distribution and cross-task generalization sampling.
Appendix~B presents more details of benchmarks and offline dataset construction.

\paragraph{Models and Baselines.}
% We evaluate our approach using three state-of-the-art open-source language models:
We build our method on three open-source language models:
% 1) \texttt{Qwen-3B} \citep{bai2023qwen}, the Qwen-3B-Instruct-v2.5 version. 2) \texttt{Mistral-7B} \citep{jiang2023mistral}, the Mistral-7B-Instruct-v0.2 version. 3) \texttt{Llama-3-8B} \citep{llama3}, the Meta-Llama-3-8B-Instruct version.
1) \texttt{Mistral-7B}~\citep{jiang2023mistral}, 2) \texttt{Gemma-7B}~\citep{team2024gemma} and 3) \texttt{Llama-3-8B}~\citep{llama3}.
We employ LoRA for parameter-efficient fine-tuning of all models.
Appendix~C presents detailed model architectures, hyperparameters, and training and evaluation setups.

% We compare GLIDER against several strong baselines:
We compare GLIDER against various strong baselines:
% 1) \texttt{ReAct}~\citep{yao2023react}, A pioneering approach that incorporates Chain-of-Thought (CoT) prompting in decision-making tasks through a structured Thought-Action-Observation loop.
1) \texttt{ReAct}~\citep{yao2023react}, a pioneering approach that incorporates CoT prompting in decision-making tasks through a structured Thought-Action-Observation loop.
% 2) \texttt{Reflexion}~\citep{shinn2023reflexion},An advanced prompt-based framework that enhances agent decision-making through verbal feedback, utilizing carefully crafted prompts to enable trajectory reflection and feedback-based replanning.
2) \texttt{Reflexion}~\citep{shinn2023reflexion}, an advanced prompt-based framework that enhances agent decision-making through self-reflective verbal feedback.
% 3) \texttt{SwiftSage}~\citep{lin2023swiftsage}, A cognitive-inspired framework based on dual-process theory, specifically designed for complex interactive reasoning and action planning tasks.
3) \texttt{SwiftSage}~\citep{lin2023swiftsage}, a dual-process cognitive framework that integrates the strengths of behavior cloning and prompting for complex interactive reasoning and action-planning tasks.
% 4) \texttt{NAT}~\citep{wang2024learning} A novel fine-tuning approach that leverages both successful and unsuccessful trajectories by incorporating quality control mechanisms to learn from failed attempts.
4) \texttt{NAT}~\citep{wang2024learning}, a fine-tuning approach that enables LLMs to learn from failure trajectories through data quality control.
% 5) \texttt{ETO}~\citep{song2024trial}, A two-phase learning framework that leverages both positive and negative trajectories. It combines behavior cloning on expert demonstrations with Direct Preference Optimization (DPO) learning from failure cases.
5) \texttt{ETO}~\citep{song2024trial}, an iterative optimization framework between exploring the environment to collect contrastive trajectory pairs and fine-tuning the policy using DPO~\citep{rafailov2023direct}.

\iffalse 
\paragraph{Training and Evaluation Setups.}
we employ LoRA for parameter-efficient fine-tuning for all language models. During the SFT phase, we train for $5$ epochs with a batch size of $32$, using the AdamW optimizer with a learning rate of $1e-4$. Policy are evaluated on both seen and unseen tasks across two benchmarks.
For the ORL phase, we train for 4 epochs with different learning rates for the actor ($1e-5$) and critic ($1e-4$) networks. The target critic network is updated using soft updates with $\tau = 0.2$, and we set the advantage weighted factor $\lambda$ to $0.99$. The training data consists of expert and medium data in a 1:2 ratio.
In the O2O phase, we assess the generalization performance by evaluating on 3 held-out tasks from ScienceWorld that were not used during training. We track the evaluation metrics during the online adaptation process and compare our hierarchical approach with several baselines, including standard Actor-Critic, AWAC, and Actor-Critic hierarchical variant. Please refer to appendix C for detailed training and evaluation setups, hyperparameters, and prompt templates for both high-level and low-level inputs.
\fi

\subsection{Primary Performance}\label{sec:performance}
\begin{table*}[!t]
\setlength{\tabcolsep}{8pt}
\renewcommand\arraystretch{0.8}
\caption{\small
\textbf{Main Results.}
% Performance comparison across three backbone models on ScienceWorld and AlfWorld benchmarks. {\small \faToggleOff} indicate prompt-based methods without model parameter update, while {\small \faToggleOn} represent fine-tuning based baselines trained using LoRA. \textcolor{red}{Red} represents the changes of GLIDER relative to the optimal results in the baselines.
Performance comparison across three backbone models on ScienceWorld and AlfWorld benchmarks. {\small \faToggleOff} indicates prompt-based methods without model parameter update, while {\small \faToggleOn} represents fine-tuning approaches using LoRA. \textcolor{red}{$\uparrow$} denotes the performance improvement of GLIDER compared to the best results among the baselines.
}
\centering
\scalebox{0.96}{
\begin{tabular}{c|l|cc|cc}
\toprule

{\multirow{2}{*}{\textbf{Backbone}}} 
&{\multirow{2}{*}{\textbf{Method}}} 
& \multicolumn{2}{c|}{\textbf{ScienceWorld}} 
& \multicolumn{2}{c}{\textbf{AlfWorld}} \\
\cmidrule{3-4}
\cmidrule{5-6}
& & Seen & Unseen & Seen & Unseen \\
\midrule

\multirow{6}{*}{\makecell{Mistral-7B}} &
{\small \faToggleOff}   ReAct       & 20.72 & 17.65 & 7.86  & 5.22  \\
& {\small \faToggleOff} Reflexion   & 21.07 & 18.11 & 11.56 & 6.00  \\
& {\small \faToggleOff} SwitchSage  & 48.40 & 45.25 & 30.29 & 26.52 \\
& {\small \faToggleOn}  NAT         & 57.12 & 50.79 & 64.43 & 68.96 \\
& {\small \faToggleOn}  ETO         & 58.17 & 51.85 & 66.84 & 71.43 \\
\cmidrule{2-6}
& {\small \faToggleOn} \textbf{\ours} 
& \textbf{67.31} {\small (\textcolor{red}{$\uparrow 15.71\%$})} 
& \textbf{65.14} {\small (\textcolor{red}{$\uparrow 25.63\%$})} 
& \textbf{70.02} {\small (\textcolor{red}{$\uparrow 4.76\%$})} 
& \textbf{74.83} {\small (\textcolor{red}{$\uparrow 4.76\%$})} \\

\midrule
\multirow{6}{*}{\makecell{Gemma-7B}} &
{\small \faToggleOff}   ReAct       & 3.58  & 3.51  & 6.43  & 2.24  \\
& {\small \faToggleOff} Reflexion   & 4.94  & 3.93  & 7.14  & 2.99  \\
& {\small \faToggleOff} SwitchSage  & 33.43 & 30.90 & 8.23  & 5.72  \\
& {\small \faToggleOn}  NAT         & 47.63 & 44.98 & 67.86 & 65.88 \\
& {\small \faToggleOn}  ETO         & 50.44 & 47.84 & 66.43 & 68.66 \\
\cmidrule{2-6}
& {\small \faToggleOn} \textbf{\ours} 
& \textbf{63.67} {\small (\textcolor{red}{$\uparrow 26.23\%$})} 
& \textbf{58.50} {\small (\textcolor{red}{$\uparrow 22.28\%$})} 
& \textbf{72.12} {\small (\textcolor{red}{$\uparrow 6.28\%$})} 
& \textbf{70.88} {\small (\textcolor{red}{$\uparrow 3.23\%$})} \\

\midrule
\multirow{6}{*}{\makecell{Llama-3-8B}} &
{\small \faToggleOff}   ReAct       & 24.76 & 22.66 & 2.86  & 3.73  \\
& {\small \faToggleOff} Reflexion   & 27.23 & 25.41 & 4.29  & 4.48  \\
& {\small \faToggleOff} SwitchSage  & 42.22 & 40.58 & 20.39 & 10.78 \\
& {\small \faToggleOn}  NAT         & 55.24 & 48.76 & 60.71 & 59.70 \\
& {\small \faToggleOn}  ETO         & 57.90 & 52.33 & 64.29 & 64.18 \\
\cmidrule{2-6}
& {\small \faToggleOn} \textbf{\ours} 
& \textbf{77.43} {\small (\textcolor{red}{$\uparrow 33.73\%$})} 
& \textbf{68.34} {\small (\textcolor{red}{$\uparrow 30.59\%$})} 
& \textbf{71.56} {\small (\textcolor{red}{$\uparrow 11.31\%$})} 
& \textbf{75.38} {\small (\textcolor{red}{$\uparrow 17.45\%$})} \\

\bottomrule
\end{tabular}
}
\label{tab:main_results}
\end{table*}

% Table.~\ref{tab:main_results} presents the comprehensive evaluation results of our proposed GLIDER compared with various baselines across three backbone models (Mistral-7B, Gemma-7B, and Llama-3-8B) on both ScienceWorld and ALFWorld benchmarks. The baselines include prompt-based methods (ReAct, Reflexion, and SwitchSage) and fine-tuning approaches (NAT and ETO).
Table~\ref{tab:main_results} illustrates the comprehensive evaluation results of GLIDER across three backbone models (Mistral-7B, Gemma-7B, and Llama-3-8B) on both ScienceWorld and ALFWorld benchmarks, compared to competent prompt-based methods (ReAct, Reflexion, and SwitchSage) and fine-tuning approaches (NAT and ETO).
% Our method consistently outperforms all baselines across different settings. 
% Notably, while fine-tuning methods (NAT and ETO) generally perform better than prompt-based approaches, GLIDER further improves upon these strong baselines.
Generally, fine-tuning approaches yield better results than prompt-based methods, and GLIDER further exceeds these strong baselines by a significant margin in both seen and unseen tasks across diverse settings.
% Specifically, with Llama-3-8B as the backbone, GLIDER achieves the better performance, reaching 77.43 (+19.53) on seen tasks and 68.34 (+16.01) on unseen tasks in ScienceWorld. 
Taking ScienceWorld as an example, GLIDER obtains the best performance with Llama-3-8B as the backbone, achieving high scores of 77.43 (+33.73\%) on seen tasks and 68.34 (+30.59\%) on unseen tasks. 
Similar improvements are observed with Mistral-7B (+15.71\% on seen tasks, +25.63\% on unseen tasks) and Gemma-7B (+26.23\% on seen tasks, +22.28\% on unseen tasks).
% Specificially, the substantial performance gains on unseen tasks (ranging from +10.66 to +16.01) demonstrate GLIDER's strong generalization capability.
Notably, the substantial performance gains on unseen tasks (ranging from +22.28\% to +30.59\%) highlight GLIDER's impressive generalization capability, which is essential for modern AI agents.
% The consistent performance boost across different model architectures and benchmarks validates the effectiveness and robustness of our approach.
In summary, the consistent performance improvement across diverse model architectures and benchmarks validates the effectiveness and robustness of our method.

\subsection{Ablation Studies}\label{sec:ablation}

% \paragraph{Ablation}
% We conduct comprehensive ablation studies to analyze the contribution of different components in GLIDER, including the hierarchical structure and training stages. 
We conduct extensive studies to analyze the respective contributions of the hierarchical structure and learning stages in GLIDER, yielding five ablations:
1) \texttt{w/ Hier(SFT)}, it ablates the offline RL stage and trains a hierarchical agent with SFT only;
2) \texttt{w/ Hier(ORL)}, it ablates the SFT stage and trains a hierarchical agent with offline RL only;
3) \texttt{w/o Hier(SFT+ORL)}, it ablates the hierarchy and trains a single-layer agent with SFT and offline RL;
4) \texttt{w/o Hier(SFT)}, it ablates the hierarchy and the offline RL stage, training a single-layer agent with SFT only;
5) \texttt{w/o Hier(ORL)}, it ablates the hierarchy and the SFT stage, training a single-layer agent with offline RL only.
%%%%%%%%%%%%%%%%%%%%%%%%%%%%%%%%%%%%%%%%%%%%%%%%%%%%%%
\begin{figure}[ht]
\centering\includegraphics[width=\columnwidth]{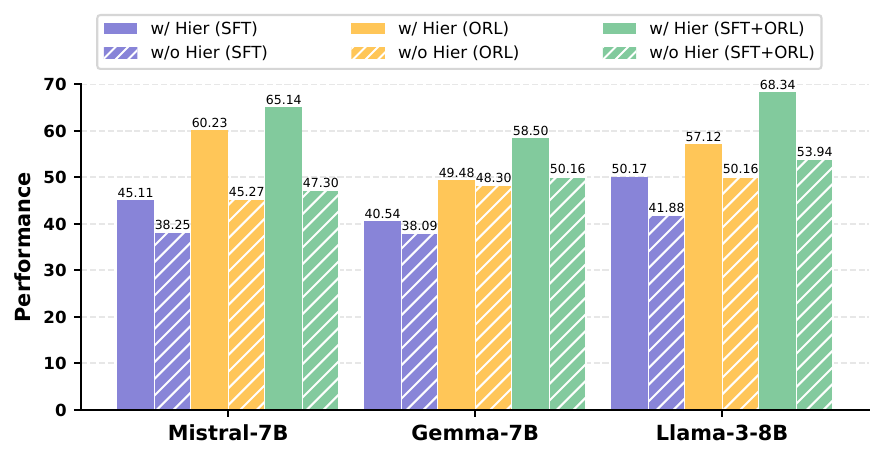}
\caption{
% Ablation study comparing different structure and training stages across three types of models. ORL w/o SFT denotes training with offline RL from scratch, while ORL w/ SFT indicates training with offline RL initialized from SFT. The w/ hierarchy and w/o hierarchy represent whether the model has hierarchical structure or not.
Ablation performance on unseen tasks in ScienceWorld across model architectures.
Solid pillars denote hierarchical models and shaded pillars indicate ablating the hierarchy.
% The purple/yellow/green pillars correspond to \textcolor{purple}{SFT}/\textcolor{yellow}{ORL}/\textcolor{green}{SFT+ORL} training stages, respectively.
The purple/yellow/green pillars correspond to SFT/ORL/SFT+ORL training stages, respectively.
}
% \vspace{-0.2cm}
\label{fig:ablation}
\end{figure}
%%%%%%%%%%%%%%%%%%%%%%%%%%%%%%%%%%%%%%%%%%%%%%%%%%%%%%

\textbf{Ablation across Model Architectures.}
% As shown in Figure~\ref{fig:ablation},we investigate the impact of hierarchical versus non-hierarchical structures and various training strategy combinations, conducting experiments across three different foundation models (Mistral-7B, Gemma-7B, and Llama-3-8B) to ensure robust findings. 
Figure~\ref{fig:ablation} presents the performance on unseen tasks in ScienceWorld.
We conduct these ablations across different language models to ensure robust findings.
% The hierarchical structure consistently improves performance across all settings, with models incorporating hierarchy outperforming their non-hierarchical counterparts by significant margins. 
First, the hierarchical structure plays a crucial part in all training stages, as models incorporating hierarchy outperform their non-hierarchical counterparts by significant margins.  
% This improvement is particularly pronounced in ORL settings, though it is also evident in SFT scenarios (e.g., $45.1$ vs $38.2$, $40.5$ vs $38.1$ and $50.2$ vs $41.9$).
The improvement is most pronounced in the full stage of SFT+ORL (green), followed by the ORL stage (yellow), and finally the SFT setting (purple).
This interesting phenomenon highlights the superiority of our method as a whole.
% Interestingly, while ORL from scratch (ORL w/o SFT, shown in yellow) underperforms SFT (shown in purple) under the same training epochs, likely due to the complexity and cost of ORL training, initializing ORL from SFT parameters (ORL w/ SFT) proves to be a more effective strategy. 
Another interesting observation is that training offline RL agents from scratch (yellow) performs better than training SFT agents (purple).
It highlights the higher potential of reinforcement fine-tuning over supervised fine-tuning, akin to the observation in DeepSeek-R1~\citep{guo2024deepseek}.
% This is likely because pure RL algorithms are generally more difficult to train compared to supervised learning.
Initializing ORL from SFT parameters (green) proves to be a more effective strategy, which is also consistent to the common practice in literature~\citep{silver2016mastering,song2024trial}.
% Moreover, different backbone models exhibit similar patterns while varying in absolute performance, with Mistral-7B and Llama-3-8B showing stronger results compared to Gemma-7B.
Moreover, using different backbones exhibits similar patterns in these ablations, while Mistral-7B and Llama-3-8B induce better performance compared to Gemma-7B.
% These results validate the effectiveness of both the hierarchical design and the staged training approach in GLIDER, with their combination yielding the best performance across all model variants.
In summary, these results validate the effectiveness of both the hierarchical structure and the multi-stage training in GLIDER, with their combination yielding the most significant results across all implementations.

% \begin{table}[tb]
% \setlength{\tabcolsep}{3pt}
% \renewcommand\arraystretch{1}
% \caption{
% % Comparison of different model architectures and sizes, evaluating the impact of hierarchical structure under SFT and ORL training settings. 
% Ablation performance on unseen tasks in ScienceWorld across model scales.
% }
% \centering
% \begin{tabular}{l|ccc|ccc}
% \toprule
% \multicolumn{1}{c|}{\multirow{2}{*}{\textbf{Model}}} & \multicolumn{3}{c|}{\textbf{w/o Hier}} & \multicolumn{3}{c}{\textbf{w/ Hier}} \\
% \cmidrule(l){2-4} \cmidrule(l){5-7}
% & SFT & ORL & SFT+ORL & SFT & ORL & SFT+ORL \\
% \midrule
% Llama-1B    &37.24     &45.31   &48.48    &44.50     &50.43     &53.62 \\
% Llama-3B    &38.19     &52.47   &56.93    &48.11     &55.98     &61.29 \\
% Llama-8B    &41.88     &50.16   &53.94    &50.17     &57.12     &68.34 \\
% \bottomrule
% \end{tabular}
% \label{tab:model_scale}
% \end{table}

\begin{table}[htb]
\setlength{\tabcolsep}{4pt}  
\renewcommand\arraystretch{1}
\caption{Ablation performance on unseen tasks in ScienceWorld across model scales.~\protect\footnote{Llama-1B and Llama-3B models refer to the Meta-Llama-3.2-1B-Instruct and Meta-Llama-3.2-3B-Instruct version, respectively.}}
\centering
\scalebox{0.85}{  
\begin{tabular}{l|ccc|ccc}
\toprule
\multicolumn{1}{c|}{\multirow{2}{*}{\textbf{Model}}} & \multicolumn{3}{c|}{\textbf{w/o Hier}} & \multicolumn{3}{c}{\textbf{w/ Hier}} \\
\cmidrule(l){2-4} \cmidrule(l){5-7}
& SFT & ORL & SFT+ORL & SFT & ORL & SFT+ORL \\
\midrule
Llama-1B    &37.24     &45.31   &48.48    &44.50     &50.43     &53.62 \\
Llama-3B    &38.19     &52.47   &56.93    &48.11     &55.98     &61.29 \\
Llama-8B    &41.88     &50.16   &53.94    &50.17     &57.12     &68.34 \\
\bottomrule
\end{tabular}}
\label{tab:model_scale}
\end{table}

% \begin{table}[tb]
% \setlength{\tabcolsep}{4pt}  
% \renewcommand\arraystretch{1}
% \caption{Ablation performance on unseen tasks in ScienceWorld across model scales}
% \centering
% \scalebox{0.85}{  
% \begin{tabular}{l|ccc|ccc}
% \toprule
% \multicolumn{1}{c|}{\multirow{2}{*}{\textbf{Model}}} & \multicolumn{3}{c|}{\textbf{w/o Hier}} & \multicolumn{3}{c}{\textbf{w/ Hier}} \\
% \cmidrule(l){2-4} \cmidrule(l){5-7}
% & SFT & ORL & SFT+ORL & SFT & ORL & SFT+ORL \\
% \midrule
% Llama-1B    &37.24     &45.31   &48.48    &44.50     &50.43     &53.62 \\
% Llama-3B    &38.19     &52.47   &56.93    &48.11     &55.98     &61.29 \\
% Llama-8B    &41.88     &50.16   &53.94    &50.17     &57.12     &68.34 \\
% \bottomrule
% \end{tabular}}
% \begin{tablenotes}
%       \small
%       \item \textit{Note:} Llama-1B and Llama-3B models refer to meta-llama/Llama-3.2-1B-Instruct and meta-llama/Llama-3.2-3B-Instruct respectively.
% \end{tablenotes}
% \label{tab:model_scale}
% \end{table}

\textbf{Ablation across Model Scales.}
% Building on our ablation studies, we further investigate the impact of model architectures and parameter scales on performance. 
Further, we investigate the impact of model scales on ablation performance.
% As shown in Table.~\ref{tab:model_scale}, we evaluate both Qwen and Llama architectures ranging from 1.5B to 3B parameters under both hierarchical and non-hierarchical settings.
Table~\ref{tab:model_scale} presents the ablation results on ScienceWorld's unseen tasks with Llama models ranging from 1B to 8B parameters, consistently demonstrating the advantages of the hierarchical structure and multi-stage training pipeline.
% across different model scales. 
% Notably, even with small parameter counts, our hierarchical approach demonstrates remarkable efficiency. 
Notably, our hierarchical approach demonstrates remarkable efficiency even with small parameter counts.
% For instance, Llama-3.2-3B with hierarchy achieves 61.29 under ORL, surpassing even larger models like Mistral-7B (58.50). 
Taking the w/ Hier (SFT+ORL) as an example, Llama-3B achieves a score of 61.29, surpassing even larger models like Mistral-7B with a score of 58.50.
% These results indicate that our hierarchical structure effectively enhances model capability without requiring larger parameter counts, making it practical for resource-constrained scenarios.
This suggests that our hierarchical structure effectively enhances agent capacity without necessitating particularly large parameter counts, making our method more practical for resource-constrained scenarios.

\subsection{Generalization Analysis via Online Fine-tuning}\label{sec:generalization}
\begin{figure}[ht]
\centering\includegraphics[width=\columnwidth]{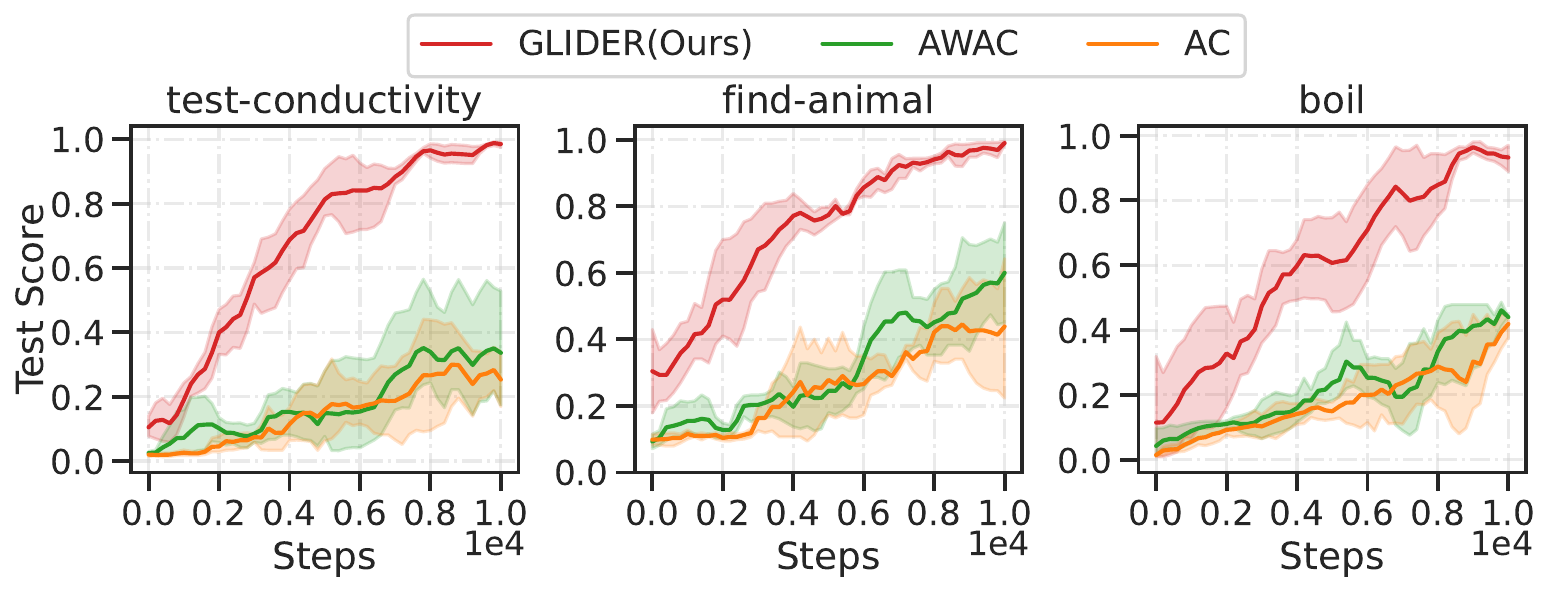}
\caption{
% Online finetuning comparison of GLIDER with AWAC and AC baselines on two ScienceWorld tasks.
Online fine-tuning performance (score/100) of GLIDER against AC and AWAC baselines in ScienceWorld.
}
\label{fig:o2o}
\end{figure}

% To evaluate the generalization capability through comparative analysis, we conduct online-to-offline (O2O) experiments with AWAC~\citep{nair2020awac} and AC~\citep{konda1999actor} as baseline methods on ScienceWorld. 
To evaluate GLIDER's generalization capacity to non-stationary environments, we test it in offline-to-online fine-tuning scenarios with comparison to the traditional AC algorithm~\citep{konda1999actor} and the classical offline-to-online algorithm AWAC~\citep{nair2020awac}.
% We categorize all experimental tasks into three distinct domains: electrical, biology, and thermodynamics. From each group, we select one representative task (test-conductivity, and find-animal, boil) for online fine-tuning while training on the remaining tasks in that group offline. 
In the ScienceWorld benchmark, we categorize the science experiment tasks into three distinct domains: electrical, biology, and thermodynamics.
From each group, we exclude one representative task (test-conductivity, find-animal, and boil) during offline training, and observe the trained agent's adaptation performance with online fine-tuning on that task.
% As shown in Figure~\ref{fig:o2o}, GLIDER exhibits strong generalization ability in two aspects. 
As shown in Figure~\ref{fig:o2o}, GLIDER exhibits strong generalization ability to new tasks in at least two aspects. 
% The higher initial test score indicates better zero-shot transfer from offline training to unseen online tasks, demonstrating effective knowledge transfer across tasks within the same domain. 
First, GLIDER achieves a higher initial test score, highlighting its superior zero-shot generalization capacity and enhanced knowledge transfer to new online tasks.
% Moreover, during the online fine-tuning phase, GLIDER shows consistently faster learning progress and achieves substantially better final performance on all tasks. 
Second, during the online fine-tuning process, GLIDER shows significantly faster adaptation and achieves substantially better final performance on all tasks.
% These results comprehensively validate GLIDER's superior generalization capability in both immediate knowledge transfer and continuous adaptation. The complete experimental setup and results for all three task groups are detailed in Appendix~B.
In summary, these results comprehensively validate GLIDER's superior generalization capability to new tasks in both zero-shot knowledge transfer and fast online adaptation, establishing GLIDER as a more competent language agent with impressive autonomous adaptability.

\subsection{Impact of Data Mixture Ratios}\label{sec:data_rate}

\begin{figure}[ht]
\centering\includegraphics[width=\columnwidth]{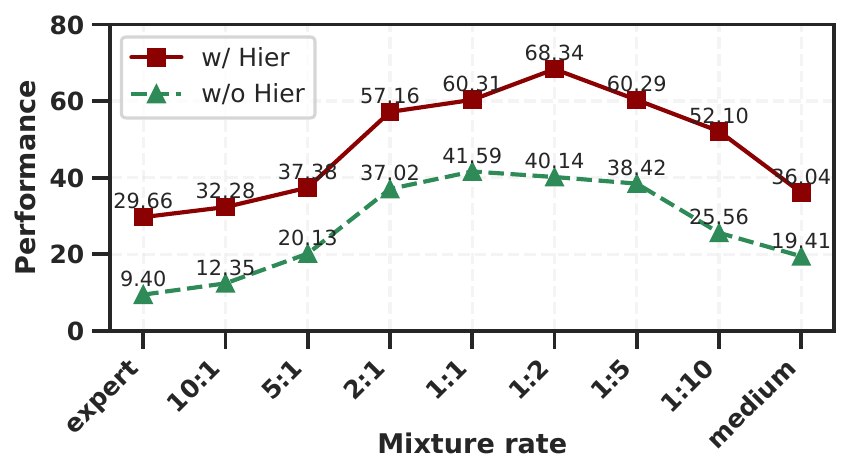}
\caption{
% Data mixture rate study showing the impact of different ratios between expert demonstrations and medium-quality data(collected from SFT training process policy) in the ORL stage.
Performance on unseen tasks in ScienceWorld with different expert-to-medium data mixture ratios in the offline RL stage with Llama-3-8B as the LLM backbone.
}
% \vspace{-0.2cm}
\label{fig:data_rate}
\end{figure}

% We investigate how different mixture ratios between expert demonstrations and medium-quality data affect model performance during offline RL training. 
We investigate how different mixture ratios between expert and medium data affect agent performance during offline RL training.
% As shown in Figure.~\ref{fig:data_rate}, the model achieves optimal performance when the expert-to-medium ratio falls within $2:1$ to $1:5$, peaking at $1:2$ ($68.3$). 
Figure~\ref{fig:data_rate} presents the GLIDER's performance (with and without hierarchy) on unseen tasks in ScienceWorld across data mixture ratios.
The agent achieves satisfactory capabilities when the expert-to-medium data ratio falls between 2:1 to 1:5, performing the best at 1:2 with a score of 68.3.
% Training with only expert demonstrations yields limited performance ($29.7$) due to insufficient trial-and-error experiences. 
% While training solely on medium-quality data performs slightly better ($36.0$), it fails to learn optimal strategies. 
An interesting phenomenon is that training with only expert demonstrations results in a limited performance of score 29.7, and training solely on medium data can obtain a slightly higher score of 36.0.
It suggests that increasing the trial-and-error experience and coverage of the state-action space (expert data is somewhat homogeneous) might facilitate the generalization performance on unseen tasks.
% A balanced mixture of both types of data is crucial for learning a comprehensive world model: expert demonstrations provide optimal behavior targets while medium-quality data contributes to environmental dynamics understanding through exploration.
Compared to supervised learning, RL naturally learns from sub-optimal data and continually reinforcing its capabilities through self-evolving.
% RL is characterized by its capacity for autonomous learning from sub-optimal data, continually reinforcing its capabilities through self-evolving.
This finding also supports our motivation to boost the LLM agent's competence via RL.
In summary, both the data quality and diversity are crucial for building capable and generalizable LLM agents.

\section{Conclusions, Limitations, and Future Work}
\label{sec:conclusion}
% In this paper, we propose GLIDER, a hierarchical framework that empowers LLMs with efficient decision-making capabilities through offline hierarchical reinforcement learning. 
We propose GLIDER, an innovative framework that empowers LLM agents with high-capacity decision-making abilities through offline hierarchical RL. 
We design a concise hierarchical model architecture that achieves superior parameter efficiency and broad applicability, efficiently grounding LLM agents to tackle complex, long-horizon tasks via sophisticated exploration and structured reasoning.
% Our experiments demonstrate GLIDER's effectiveness in handling complex, long-horizon tasks while maintaining strong generalization ability and parameter efficiency.
Extensive experiments validate GLIDER's consistent improvement on learning performance and generalization capability.

% Although our method employs a multi-stage training pipeline (SFT and ORL) that involves relatively complex training procedures, this framework's potential extends beyond decision-making tasks, as many LLM tasks can be reformulated as sequential decision-making processes through PRM. For future work, we aim to streamline the training pipeline while maintaining its efficiency advantage, inspired by DeepSeek-R1's recent advances in RL training. Moreover, we plan to extend our method to more domains such as mathematical reasoning and code generation tasks, leveraging the framework's inherent capability in handling sequential decision-making problems.
Though, our method employs a multi-stage pipeline that involves a somewhat complex training procedure.
A promising future work is to streamline the training pipeline while maintaining high efficiency, inspired by DeepSeek-R1's recent advances in reinforcement fine-tuning.
Further, our framework's potential can extend beyond strict agent tasks, since many LLM tasks can also be reformulated as the sequential decision-making paradigm through process reward model (PRM).
A crucial future step is to extend our method to broader domains such as mathematical reasoning and code generation tasks, unleashing the inherent capability of hierarchical agents in addressing complicated problems.

\clearpage 
\section*{Acknowledgements}
We thank Xuyang Hu and Zhenhong Sun for their helpful discussions. This work was supported in part by the National Natural Science Foundation of China (Nos. 62376122 and 72394363), in part by the AI \& AI for Science Project of Nanjing University, and in part by the Nanjing University Integrated Research Platform of the Ministry of Education-Top Talents Program.

\section*{Impact Statement}
This paper presents work whose goal is to advance the field of Machine Learning. There are many potential societal consequences of our work, none which we feel must be specifically highlighted here.

\bibliography{glider}

\begin{thebibliography}{63}
\providecommand{\natexlab}[1]{#1}
\providecommand{\url}[1]{\texttt{#1}}
\expandafter\ifx\csname urlstyle\endcsname\relax
  \providecommand{\doi}[1]{doi: #1}\else
  \providecommand{\doi}{doi: \begingroup \urlstyle{rm}\Url}\fi

\bibitem[Ahn et~al.(2022)Ahn, Brohan, Brown, Chebotar, Cortes, David, Finn, Fu, Gopalakrishnan, Hausman, et~al.]{ahn2022can}
Ahn, M., Brohan, A., Brown, N., Chebotar, Y., Cortes, O., David, B., Finn, C., Fu, C., Gopalakrishnan, K., Hausman, K., et~al.
\newblock Do as {I} can, not as {I} say: Grounding language in robotic affordances.
\newblock In \emph{Proceedings of Conference on Robot Learning}, 2022.

\bibitem[Bacon et~al.(2017)Bacon, Harb, and Precup]{bacon2017option}
Bacon, P.-L., Harb, J., and Precup, D.
\newblock The option-critic architecture.
\newblock In \emph{Proceedings of AAAI Conference on Artificial Intelligence}, volume~31, 2017.

\bibitem[Black et~al.(2024)Black, Brown, Driess, Esmail, Equi, Finn, Fusai, Groom, Hausman, Ichter, et~al.]{black2024pi}
Black, K., Brown, N., Driess, D., Esmail, A., Equi, M., Finn, C., Fusai, N., Groom, L., Hausman, K., Ichter, B., et~al.
\newblock $\pi_0$: A vision-language-action flow model for general robot control.
\newblock \emph{arXiv preprint arXiv:2410.24164}, 2024.

\bibitem[Brown et~al.(2020)Brown, Mann, Ryder, Subbiah, Kaplan, et~al.]{brown2020language}
Brown, T., Mann, B., Ryder, N., Subbiah, M., Kaplan, J.~D., et~al.
\newblock Language models are few-shot learners.
\newblock In \emph{Advances in Neural Information Processing Systems}, volume~33, pp.\  1877--1901, 2020.

\bibitem[Burda et~al.(2019)Burda, Edwards, Storkey, and Klimov]{burda2019exploration}
Burda, Y., Edwards, H., Storkey, A., and Klimov, O.
\newblock Exploration by random network distillation.
\newblock In \emph{Proceedings of International Conference on Learning Representations}, 2019.

\bibitem[Chen et~al.(2023)Chen, Shu, Shareghi, Collier, Narasimhan, and Yao]{chen2023fireact}
Chen, B., Shu, C., Shareghi, E., Collier, N., Narasimhan, K., and Yao, S.
\newblock {FireAct}: Toward language agent fine-tuning.
\newblock \emph{arXiv preprint arXiv:2310.05915}, 2023.

\bibitem[Chen et~al.(2024)Chen, Su, Zuo, Yang, Yuan, Chan, Yu, Lu, Hung, Qian, et~al.]{chen2024agentverse}
Chen, W., Su, Y., Zuo, J., Yang, C., Yuan, C., Chan, C.-M., Yu, H., Lu, Y., Hung, Y.-H., Qian, C., et~al.
\newblock {AgentVerse}: Facilitating multi-agent collaboration and exploring emergent behaviors.
\newblock In \emph{Proceedings of International Conference on Learning Representations}, 2024.

\bibitem[Dietterich(2000)]{dietterich2000hierarchical}
Dietterich, T.~G.
\newblock Hierarchical reinforcement learning with the maxq value function decomposition.
\newblock \emph{Journal of Artificial Intelligence Research}, 13:\penalty0 227--303, 2000.

\bibitem[Dwaracherla et~al.(2024)Dwaracherla, Asghari, Hao, and Van~Roy]{dwaracherla2024efficient}
Dwaracherla, V., Asghari, S.~M., Hao, B., and Van~Roy, B.
\newblock Efficient exploration for llms.
\newblock In \emph{Proceedings of International Conference on Machine Learning}, pp.\  12215–12227, 2024.

\bibitem[Fujimoto et~al.(2019)Fujimoto, Meger, and Precup]{fujimoto2019off}
Fujimoto, S., Meger, D., and Precup, D.
\newblock Off-policy deep reinforcement learning without exploration.
\newblock In \emph{Proceedings of International Conference on Machine Learning}, pp.\  2052--2062, 2019.

\bibitem[Gandhi et~al.(2023)Gandhi, Sadigh, and Goodman]{gandhi2023strategic}
Gandhi, K., Sadigh, D., and Goodman, N.~D.
\newblock Strategic reasoning with language models.
\newblock \emph{arXiv preprint arXiv:2305.19165}, 2023.

\bibitem[Guo et~al.(2024)Guo, Zhu, Yang, Xie, Dong, Zhang, Chen, Bi, Wu, Li, et~al.]{guo2024deepseek}
Guo, D., Zhu, Q., Yang, D., Xie, Z., Dong, K., Zhang, W., Chen, G., Bi, X., Wu, Y., Li, Y., et~al.
\newblock Deepseek-coder: When the large language model meets programming--the rise of code intelligence.
\newblock \emph{arXiv preprint arXiv:2401.14196}, 2024.

\bibitem[Guo et~al.(2025)Guo, Yang, Zhang, Song, Zhang, et~al.]{guo2025deepseek}
Guo, D., Yang, D., Zhang, H., Song, J., Zhang, R., et~al.
\newblock {DeepSeek-R1}: Incentivizing reasoning capability in llms via reinforcement learning.
\newblock \emph{arXiv preprint arXiv:2501.12948}, 2025.

\bibitem[Haarnoja et~al.(2018)Haarnoja, Zhou, Abbeel, and Levine]{haarnoja2018soft}
Haarnoja, T., Zhou, A., Abbeel, P., and Levine, S.
\newblock Soft actor-critic: Off-policy maximum entropy deep reinforcement learning with a stochastic actor.
\newblock In \emph{Proceedings of International Conference on Machine Learning}, pp.\  1861--1870, 2018.

\bibitem[Hu et~al.(2022)Hu, Wallis, Allen-Zhu, Li, Wang, et~al.]{hu2022lora}
Hu, E.~J., Wallis, P., Allen-Zhu, Z., Li, Y., Wang, S., et~al.
\newblock {LoRA}: Low-rank adaptation of large language models.
\newblock In \emph{Proceedings of International Conference on Learning Representations}, 2022.

\bibitem[Jiang et~al.(2023)Jiang, Sablayrolles, Mensch, Bamford, Chaplot, Casas, Bressand, Lengyel, Lample, Saulnier, et~al.]{jiang2023mistral}
Jiang, A.~Q., Sablayrolles, A., Mensch, A., Bamford, C., Chaplot, D.~S., Casas, D. d.~l., Bressand, F., Lengyel, G., Lample, G., Saulnier, L., et~al.
\newblock Mistral 7b.
\newblock \emph{arXiv preprint arXiv:2310.06825}, 2023.

\bibitem[Konda \& Tsitsiklis(1999)Konda and Tsitsiklis]{konda1999actor}
Konda, V. and Tsitsiklis, J.
\newblock Actor-critic algorithms.
\newblock In \emph{Advances in Neural Information Processing Systems}, volume~12, pp.\  75993--76005, 1999.

\bibitem[Kostrikov et~al.(2021)Kostrikov, Fergus, Tompson, and Nachum]{kostrikov2021offline}
Kostrikov, I., Fergus, R., Tompson, J., and Nachum, O.
\newblock Offline reinforcement learning with fisher divergence critic regularization.
\newblock In \emph{International Conference on Machine Learning}, pp.\  5774--5783, 2021.

\bibitem[Kostrikov et~al.(2022)Kostrikov, Nair, and Levine]{kostrikov2022offline}
Kostrikov, I., Nair, A., and Levine, S.
\newblock Offline reinforcement learning with implicit {Q}-learning.
\newblock In \emph{Proceedings of International Conference on Learning Representations}, 2022.

\bibitem[Kumar et~al.(2020)Kumar, Zhou, Tucker, and Levine]{kumar2020conservative}
Kumar, A., Zhou, A., Tucker, G., and Levine, S.
\newblock Conservative q-learning for offline reinforcement learning.
\newblock In \emph{Advances in Neural Information Processing Systems}, volume~33, pp.\  1179--1191, 2020.

\bibitem[Lee et~al.(2022)Lee, Seo, Lee, Abbeel, and Shin]{lee2022offline}
Lee, S., Seo, Y., Lee, K., Abbeel, P., and Shin, J.
\newblock Offline-to-online reinforcement learning via balanced replay and pessimistic {Q}-ensemble.
\newblock In \emph{Proceedings of Conference on Robot Learning}, pp.\  1702--1712, 2022.

\bibitem[Levine et~al.(2020)Levine, Kumar, Tucker, and Fu]{levine2020offline}
Levine, S., Kumar, A., Tucker, G., and Fu, J.
\newblock Offline reinforcement learning: Tutorial, review, and perspectives on open problems.
\newblock \emph{arXiv preprint arXiv:2005.01643}, 2020.

\bibitem[Levy et~al.(2019)Levy, Konidaris, Platt, and Saenko]{levy2019learning}
Levy, A., Konidaris, G., Platt, R., and Saenko, K.
\newblock Learning multi-level hierarchies with hindsight.
\newblock In \emph{Proceedings of International Conference on Learning Representations}, 2019.

\bibitem[Li et~al.(2020)Li, Florensa, Clavera, and Abbeel]{hippo2020}
Li, A., Florensa, C., Clavera, I., and Abbeel, P.
\newblock Sub-policy adaptation for hierarchical reinforcement learning.
\newblock In \emph{Proceedings of International Conference on Learning Representations}, 2020.

\bibitem[Li et~al.(2022)Li, Puig, Paxton, Du, Wang, Fan, Chen, Huang, Aky{\"u}rek, Anandkumar, et~al.]{li2022pre}
Li, S., Puig, X., Paxton, C., Du, Y., Wang, C., Fan, L., Chen, T., Huang, D.-A., Aky{\"u}rek, E., Anandkumar, A., et~al.
\newblock Pre-trained language models for interactive decision-making.
\newblock In \emph{Advances in Neural Information Processing Systems}, volume~35, pp.\  31199--31212, 2022.

\bibitem[Li et~al.(2024)Li, Xie, Shao, Chen, Jiang, and Nie]{li2024optimus}
Li, Z., Xie, Y., Shao, R., Chen, G., Jiang, D., and Nie, L.
\newblock Optimus-1: Hybrid multimodal memory empowered agents excel in long-horizon tasks.
\newblock In \emph{Advances in Neural Information Processing Systems}, volume~37, 2024.

\bibitem[Lin et~al.(2023)Lin, Fu, Yang, Brahman, Huang, Bhagavatula, Ammanabrolu, Choi, and Ren]{lin2023swiftsage}
Lin, B.~Y., Fu, Y., Yang, K., Brahman, F., Huang, S., Bhagavatula, C., Ammanabrolu, P., Choi, Y., and Ren, X.
\newblock {SwiftSage}: a generative agent with fast and slow thinking for complex interactive tasks.
\newblock In \emph{Advances in Neural Information Processing Systems}, pp.\  23813--23825, 2023.

\bibitem[Luo et~al.(2024)Luo, Xu, Zhao, Sun, Geng, Hu, Tao, Ma, Lin, and Jiang]{luo2024wizardcoder}
Luo, Z., Xu, C., Zhao, P., Sun, Q., Geng, X., Hu, W., Tao, C., Ma, J., Lin, Q., and Jiang, D.
\newblock {WizardCoder}: Empowering code large language models with evol-instruct.
\newblock In \emph{Proceedings of International Conference on Learning Representations}, 2024.

\bibitem[Ma et~al.(2024)Ma, Mi, Zeng, Yan, Wu, Lin, Zhang, and Wang]{ma2024large}
Ma, W., Mi, Q., Zeng, Y., Yan, X., Wu, Y., Lin, R., Zhang, H., and Wang, J.
\newblock Large language models play {StarCraft II}: Benchmarks and a chain of summarization approach.
\newblock In \emph{Advances in Neural Information Processing Systems}, volume~37, 2024.

\bibitem[Mahankali et~al.(2024)Mahankali, Hong, Sekhari, Rakhlin, and Agrawal]{mahankali2024random}
Mahankali, S., Hong, Z.-W., Sekhari, A., Rakhlin, A., and Agrawal, P.
\newblock Random latent exploration for deep reinforcement learning.
\newblock In \emph{Proceedings of International Conference on Machine Learning}, pp.\  34219–34252, 2024.

\bibitem[Meta(2024)]{llama3}
Meta.
\newblock {Introducing Meta Llama 3}: The most capable openly available {LLM} to date, 2024.
\newblock \url{https://ai.meta.com/blog/meta-llama-3/}.

\bibitem[Nachum et~al.(2018)Nachum, Gu, Lee, and Levine]{nachum2018data}
Nachum, O., Gu, S.~S., Lee, H., and Levine, S.
\newblock Data-efficient hierarchical reinforcement learning.
\newblock In \emph{Advances in Neural Information Processing Systems}, volume~31, pp.\  3307–3317, 2018.

\bibitem[Nair et~al.(2020)Nair, Gupta, Dalal, and Levine]{nair2020awac}
Nair, A., Gupta, A., Dalal, M., and Levine, S.
\newblock {AWAC}: Accelerating online reinforcement learning with offline datasets.
\newblock \emph{arXiv preprint arXiv:2006.09359}, 2020.

\bibitem[Ouyang et~al.(2022)Ouyang, Wu, Jiang, Almeida, Wainwright, Mishkin, Zhang, Agarwal, Slama, Ray, et~al.]{ouyang2022training}
Ouyang, L., Wu, J., Jiang, X., Almeida, D., Wainwright, C., Mishkin, P., Zhang, C., Agarwal, S., Slama, K., Ray, A., et~al.
\newblock Training language models to follow instructions with human feedback.
\newblock In \emph{Advances in Neural Information Processing Systems}, volume~35, pp.\  27730--27744, 2022.

\bibitem[Qiao et~al.(2024)Qiao, Fang, Zhang, Zhu, Chen, Deng, Jiang, Xie, Huang, and Chen]{qiao2024agent}
Qiao, S., Fang, R., Zhang, N., Zhu, Y., Chen, X., Deng, S., Jiang, Y., Xie, P., Huang, F., and Chen, H.
\newblock Agent planning with world knowledge model.
\newblock In \emph{Advances in Neural Information Processing Systems}, volume~37, 2024.

\bibitem[Rafailov et~al.(2023)Rafailov, Sharma, Mitchell, Manning, Ermon, and Finn]{rafailov2023direct}
Rafailov, R., Sharma, A., Mitchell, E., Manning, C.~D., Ermon, S., and Finn, C.
\newblock Direct preference optimization: Your language model is secretly a reward model.
\newblock In \emph{Advances in Neural Information Processing Systems}, volume~36, pp.\  53728–53741, 2023.

\bibitem[Rocamonde et~al.(2024)Rocamonde, Montesinos, Nava, Perez, and Lindner]{rocamonde2024vision}
Rocamonde, J., Montesinos, V., Nava, E., Perez, E., and Lindner, D.
\newblock Vision-language models are zero-shot reward models for reinforcement learning.
\newblock In \emph{Proceedings of International Conference on Learning Representations}, 2024.

\bibitem[Shah et~al.(2023)Shah, Osi{\'n}ski, Levine, et~al.]{shah2023lm}
Shah, D., Osi{\'n}ski, B., Levine, S., et~al.
\newblock {LM-Nav}: Robotic navigation with large pre-trained models of language, vision, and action.
\newblock In \emph{Proceedings of Conference on Robot Learning}, pp.\  492--504, 2023.

\bibitem[Shinn et~al.(2023)Shinn, Cassano, Gopinath, Narasimhan, and Yao]{shinn2023reflexion}
Shinn, N., Cassano, F., Gopinath, A., Narasimhan, K.~R., and Yao, S.
\newblock Reflexion: language agents with verbal reinforcement learning.
\newblock In \emph{Advances in Neural Information Processing Systems}, volume~36, pp.\  8634–8652, 2023.

\bibitem[Shridhar et~al.(2021)Shridhar, Yuan, Cote, Bisk, Trischler, and Hausknecht]{shridhar2021alfworld}
Shridhar, M., Yuan, X., Cote, M.-A., Bisk, Y., Trischler, A., and Hausknecht, M.
\newblock {ALFWorld}: Aligning text and embodied environments for interactive learning.
\newblock In \emph{Proceedings of International Conference on Learning Representations}, 2021.

\bibitem[Silver et~al.(2016)Silver, Huang, Maddison, Guez, Sifre, et~al.]{silver2016mastering}
Silver, D., Huang, A., Maddison, C.~J., Guez, A., Sifre, L., et~al.
\newblock Mastering the game of {Go} with deep neural networks and tree search.
\newblock \emph{Nature}, 529\penalty0 (7587):\penalty0 484--489, 2016.

\bibitem[Silver et~al.(2021)Silver, Singh, Precup, and Sutton]{silver2021reward}
Silver, D., Singh, S., Precup, D., and Sutton, R.~S.
\newblock Reward is enough.
\newblock \emph{Artificial Intelligence}, 299:\penalty0 103535, 2021.

\bibitem[Snell et~al.(2023)Snell, Kostrikov, Su, Yang, and Levine]{snell2023offline}
Snell, C.~V., Kostrikov, I., Su, Y., Yang, S., and Levine, S.
\newblock Offline {RL} for natural language generation with implicit language {Q}-learning.
\newblock In \emph{Proceedings of International Conference on Learning Representations}, 2023.

\bibitem[Song et~al.(2024)Song, Yin, Yue, Huang, Li, and Lin]{song2024trial}
Song, Y., Yin, D., Yue, X., Huang, J., Li, S., and Lin, B.~Y.
\newblock Trial and error: Exploration-based trajectory optimization for llm agents.
\newblock In \emph{Proceedings of Annual Meeting of the Association for Computational Linguistics}, 2024.

\bibitem[Sutton et~al.(1999)Sutton, Precup, and Singh]{sutton1999between}
Sutton, R.~S., Precup, D., and Singh, S.
\newblock Between mdps and semi-mdps: A framework for temporal abstraction in reinforcement learning.
\newblock \emph{Artificial intelligence}, 112\penalty0 (1-2):\penalty0 181--211, 1999.

\bibitem[Szot et~al.(2024)Szot, Schwarzer, Agrawal, Mazoure, Metcalf, Talbott, Mackraz, Hjelm, and Toshev]{szot2024large}
Szot, A., Schwarzer, M., Agrawal, H., Mazoure, B., Metcalf, R., Talbott, W., Mackraz, N., Hjelm, R.~D., and Toshev, A.~T.
\newblock Large language models as generalizable policies for embodied tasks.
\newblock In \emph{Proceedings of International Conference on Learning Representations}, 2024.

\bibitem[Tan et~al.(2024)Tan, Zhang, Liu, Zheng, Wang, and An]{tan2024true}
Tan, W., Zhang, W., Liu, S., Zheng, L., Wang, X., and An, B.
\newblock True knowledge comes from practice: Aligning large language models with embodied environments via reinforcement learning.
\newblock In \emph{Proceedings of International Conference on Learning Representations}, 2024.

\bibitem[Team et~al.(2024)Team, Mesnard, Hardin, Dadashi, Bhupatiraju, Pathak, Sifre, Rivi{\`e}re, Kale, Love, et~al.]{team2024gemma}
Team, G., Mesnard, T., Hardin, C., Dadashi, R., Bhupatiraju, S., Pathak, S., Sifre, L., Rivi{\`e}re, M., Kale, M.~S., Love, J., et~al.
\newblock Gemma: Open models based on gemini research and technology.
\newblock \emph{arXiv preprint arXiv:2403.08295}, 2024.

\bibitem[Wang et~al.(2023{\natexlab{a}})Wang, Xu, Lan, Hu, Lan, Lee, and Lim]{wang2023plan}
Wang, L., Xu, W., Lan, Y., Hu, Z., Lan, Y., Lee, R. K.-W., and Lim, E.-P.
\newblock Plan-and-solve prompting: Improving zero-shot chain-of-thought reasoning by large language models.
\newblock In \emph{Proceedings of Annual Meeting of the Association for Computational Linguistics}, pp.\  9--14, 2023{\natexlab{a}}.

\bibitem[Wang et~al.(2022)Wang, Jansen, C{\^o}t{\'e}, and Ammanabrolu]{wang2022scienceworld}
Wang, R., Jansen, P., C{\^o}t{\'e}, M.-A., and Ammanabrolu, P.
\newblock {ScienceWorld}: Is your agent smarter than a 5th grader?
\newblock In \emph{Proceedings of Empirical Methods in Natural Language Processing}, pp.\  11279--11298, 2022.

\bibitem[Wang et~al.(2024)Wang, Li, Han, Zhang, and Baldwin]{wang2024learning}
Wang, R., Li, H., Han, X., Zhang, Y., and Baldwin, T.
\newblock Learning from failure: Integrating negative examples when fine-tuning large language models as agents.
\newblock \emph{arXiv preprint arXiv:2402.11651}, 2024.

\bibitem[Wang et~al.(2023{\natexlab{b}})Wang, Yang, Gao, Lin, Chen, Wu, Jia, Song, and Huang]{wang2023train}
Wang, S., Yang, Q., Gao, J., Lin, M., Chen, H., Wu, L., Jia, N., Song, S., and Huang, G.
\newblock Train once, get a family: State-adaptive balances for offline-to-online reinforcement learning.
\newblock In \emph{Advances in Neural Information Processing Systems}, volume~36, pp.\  47081–47104, 2023{\natexlab{b}}.

\bibitem[Wei et~al.(2022)Wei, Wang, Schuurmans, Bosma, Xia, Chi, Le, Zhou, et~al.]{wei2022chain}
Wei, J., Wang, X., Schuurmans, D., Bosma, M., Xia, F., Chi, E., Le, Q.~V., Zhou, D., et~al.
\newblock Chain-of-thought prompting elicits reasoning in large language models.
\newblock In \emph{Advances in Neural Information Processing Systems}, volume~35, pp.\  24824--24837, 2022.

\bibitem[Wooldridge \& Jennings(1995)Wooldridge and Jennings]{wooldridge1995intelligent}
Wooldridge, M. and Jennings, N.~R.
\newblock Intelligent agents: Theory and practice.
\newblock \emph{The Knowledge Engineering Review}, 10\penalty0 (2):\penalty0 115--152, 1995.

\bibitem[Xi et~al.(2023)Xi, Chen, Guo, He, Ding, et~al.]{xi2023rise}
Xi, Z., Chen, W., Guo, X., He, W., Ding, Y., et~al.
\newblock The rise and potential of large language model based agents: A survey.
\newblock \emph{arXiv preprint arXiv:2309.07864}, 2023.

\bibitem[Xu et~al.(2024)Xu, Yu, Fang, Wang, and Wu]{xu2024language}
Xu, Z., Yu, C., Fang, F., Wang, Y., and Wu, Y.
\newblock Language agents with reinforcement learning for strategic play in the werewolf game.
\newblock In \emph{Proceedings of International Conference on Machine Learning}, pp.\  55434–55464, 2024.

\bibitem[Yao et~al.(2023{\natexlab{a}})Yao, Yu, Zhao, Shafran, Griffiths, Cao, and Narasimhan]{yao2023tree}
Yao, S., Yu, D., Zhao, J., Shafran, I., Griffiths, T.~L., Cao, Y., and Narasimhan, K.~R.
\newblock Tree of thoughts: Deliberate problem solving with large language models.
\newblock In \emph{Advances in Neural Information Processing Systems}, volume~36, pp.\  11809–11822, 2023{\natexlab{a}}.

\bibitem[Yao et~al.(2023{\natexlab{b}})Yao, Zhao, Yu, Du, Shafran, Narasimhan, and Cao]{yao2023react}
Yao, S., Zhao, J., Yu, D., Du, N., Shafran, I., Narasimhan, K., and Cao, Y.
\newblock {ReAct}: Synergizing reasoning and acting in language models.
\newblock In \emph{Proceedings of International Conference on Learning Representations}, 2023{\natexlab{b}}.

\bibitem[Yu \& Zhang(2023)Yu and Zhang]{yu2023actor}
Yu, Z. and Zhang, X.
\newblock Actor-critic alignment for offline-to-online reinforcement learning.
\newblock In \emph{Proceedings of the 40th International Conference on Machine Learning}, pp.\  40452--40474, 2023.

\bibitem[Zeng et~al.(2023)Zeng, Liu, Lu, Wang, Liu, Dong, and Tang]{zeng2023agenttuning}
Zeng, A., Liu, M., Lu, R., Wang, B., Liu, X., Dong, Y., and Tang, J.
\newblock {AgentTuning}: Enabling generalized agent abilities for {LLMs}.
\newblock \emph{arXiv preprint arXiv:2310.12823}, 2023.

\bibitem[Zhai et~al.(2024)Zhai, Bai, Lin, Pan, Tong, Zhou, Suhr, Xie, LeCun, Ma, et~al.]{zhai2024fine}
Zhai, Y., Bai, H., Lin, Z., Pan, J., Tong, S., Zhou, Y., Suhr, A., Xie, S., LeCun, Y., Ma, Y., et~al.
\newblock Fine-tuning large vision-language models as decision-making agents via reinforcement learning.
\newblock In \emph{Advances in Neural Information Processing Systems}, volume~37, 2024.

\bibitem[Zhou et~al.(2023)Zhou, Sch{\"a}rli, Hou, Wei, Scales, Wang, Schuurmans, Cui, Bousquet, Le, and Chi]{zhou2023leasttomost}
Zhou, D., Sch{\"a}rli, N., Hou, L., Wei, J., Scales, N., Wang, X., Schuurmans, D., Cui, C., Bousquet, O., Le, Q.~V., and Chi, E.~H.
\newblock Least-to-most prompting enables complex reasoning in large language models.
\newblock In \emph{Proceedings of International Conference on Learning Representations}, 2023.

\bibitem[Zhou et~al.(2024)Zhou, Zanette, Pan, Levine, and Kumar]{zhou2024archer}
Zhou, Y., Zanette, A., Pan, J., Levine, S., and Kumar, A.
\newblock {ArCHer}: Training language model agents via hierarchical multi-turn rl.
\newblock In \emph{Proceedings of International Conference on Machine Learning}, pp.\  62178–62209, 2024.

\end{thebibliography}
\bibliographystyle{icml2025}

% APPENDIX
\appendix
\onecolumn
\section*{Appendix A. Algorithm Pseudocodes}
Based on the implementations in Section~\ref{sec:method}, we summarize the brief procedure of GLIDER. Algorithm~\ref{algo} presents the complete training pipeline of GLIDER, which consists of three stages. In SFT stage, we perform behavioral cloning to train both high-level and low-level policies using demonstration data. Notably, both policies share the same LLM parameters but are differentiated through distinct prompts, which significantly reduces the parameter count while maintaining the hierarchical structure. The high-level policy prompt focuses on task decomposition, while the low-level policy prompt emphasizes primitive action generation. In ORL Stage, where both policies and critics are updated using data from high-level ($\mathcal{B}_\mathcal{H}$) and low-level ($\mathcal{B}_\mathcal{L}$) replay buffers, where contains a balanced mixture of expert demonstrations and medium-quality trajectories. Critics are updated through bootstrapping, while policies are optimized via a policy gradient. O2O stage describes the optional online adaptation process. Due to parameter sharing between policies, we cannot strictly fix the low-level policy parameters. Instead, we maintain low-level policy performance by continually training on offline demonstration data while simultaneously fine-tuning the high-level policy using newly collected transition data.

\begin{algorithm*}[!h]
\caption{GLIDER: A Hierarchical Framework for LLM-based Decision Making}
\label{algo}
\KwIn{
    \begin{itemize}[noitemsep,leftmargin=*]
        \item High-level replay buffer: $\mathcal{B_H}=\{(d,o_0,g_0,R_0,o_c...,o_{t},g_{t},R_t,o_{t+c})^{(i)}\}$, where $R_t=\sum_{i=t}^{t+c-1}r_i$
        \item Low-level replay buffer: $\mathcal{B_L}=\{(g_t,o_t,a_t,\hat{r}_t,o_{t+1},...,o_{t+c-1},a_{t+c-1},\hat{r}_{t+c-1},o_{t+c})^{(j)}\}$
        \item Environment: env
        \item Hierarchical policy: $\pi^h_\theta,\pi^l_\theta$
        \item Hierarchical critic: $Q_\phi^h, V_\psi^h, Q_\phi^l, V_\psi^l$
        \item Hyperparameters: discount $\gamma$, update rate $\tau$, regularization weight $\lambda$
    \end{itemize}
}
\KwOut{Optimized hierarchical policy $\pi_\theta=\{\pi^h_\theta,\pi^l_\theta\}$}
\textcolor{blue}{// Stage 1: SFT}\\
\For{iteration $i=1,2,...$}{
    Update hierarchical policy via BC loss (Eq.\ref{bc_loss},Eq.\ref{policy_prob})
}
\textcolor{blue}{// Stage 2: ORL} \\
\For{iteration $i=1,2,...$}{
    Sample batches from $\mathcal{B_H}$ and $\mathcal{B_L}$ \\
    Update critics via bootstrapping (Eq.\ref{q_value}, Eq.\ref{value}) \\
    Update policies via policy gradient (Eq.\ref{awac}) \\
    Soft update target networks: $\bar{\eta} \leftarrow (1 - \tau)\bar{\eta} + \tau\eta$
}
\textcolor{blue}{// Stage 3: O2O (Optional)}
Fix low-level policy $\pi_\theta^l$\\
\For{episode $=1,2,...$}{
    $o_0 \leftarrow$ env.reset(task), $d \leftarrow$ env.task$\_$description() \\
    Initialize trajectory $\xi \leftarrow \varnothing$ \\
    \For{$t=1,...,T$}{
        Sample subtask: $g_t \sim \pi_\theta^h(\cdot \mid d,o_t)$ \\
        \For{step $h=1,...,c$}{
            Sample action: $a_t \sim \pi_\theta^l(\cdot \mid g_t,o_t)$ \\
            $r_t, o_{t+1}, done \leftarrow$ env.step($a_t$)
        }
        Store transition: $\xi \leftarrow \xi \cup (d,o_t,g_t,R_t,o_t+c)$ \\
        $t \leftarrow t + c$
    }
    Store trajectory: $\mathcal{B_H} \leftarrow \mathcal{B_H} \cup \{\xi\}$\\
    Update high-level policy and critic using $\mathcal{B_H}$ (Eq.\ref{q_value}-\ref{awac})
}
\end{algorithm*}

\clearpage
\section*{Appendix B. Dataset Information}
\subsection*{Benchmarks.} 
We evaluate GLIDER on two popular language-based interactive decision-making tasks:
\begin{enumerate}
\item \textbf{ScienceWorld}~\citep{wang2022scienceworld} is a textual environment for elementary science experiments, featuring 30 tasks across 10 categories. 
Agents must demonstrate scientific understanding through interactive experimentation, with progress measured by a dense reward (0 to 1) at each step.
\item \textbf{ALFWorld}~\citep{shridhar2021alfworld} simulates household environments that require navigation and object manipulation in a sparse, binary reward setting.
The reward is 1 only upon successful task completion, and 0 otherwise.
\end{enumerate}
Beyond standard evaluation on seen tasks, it includes unseen scenarios to assess generalization ability. Table~\ref{tab:dataset_statistics} presents the statistical information of our datasets. Both ScienceWorld and ALFWorld contain Text-Seen and Text-Unseen test sets, where Text-Unseen comprises out-of-distribution variations to evaluate the generalization capabilities of different agents.
\begin{table}[htb]
\centering
\caption{Dataset statistics.}
\begin{tabular}{lccc}
\toprule
\textbf{Dataset} & \textbf{Train} & \textbf{Text-Seen} & \textbf{Text-Unseen} \\
\midrule
ScienceWorld & 1,483 & 194 & 211 \\
ALFWorld & 3,119 & 140 & 134 \\
\bottomrule
\end{tabular}
\label{tab:dataset_statistics}
\end{table}

\subsection*{Expert Demonstration}
\begin{figure}[!htp]
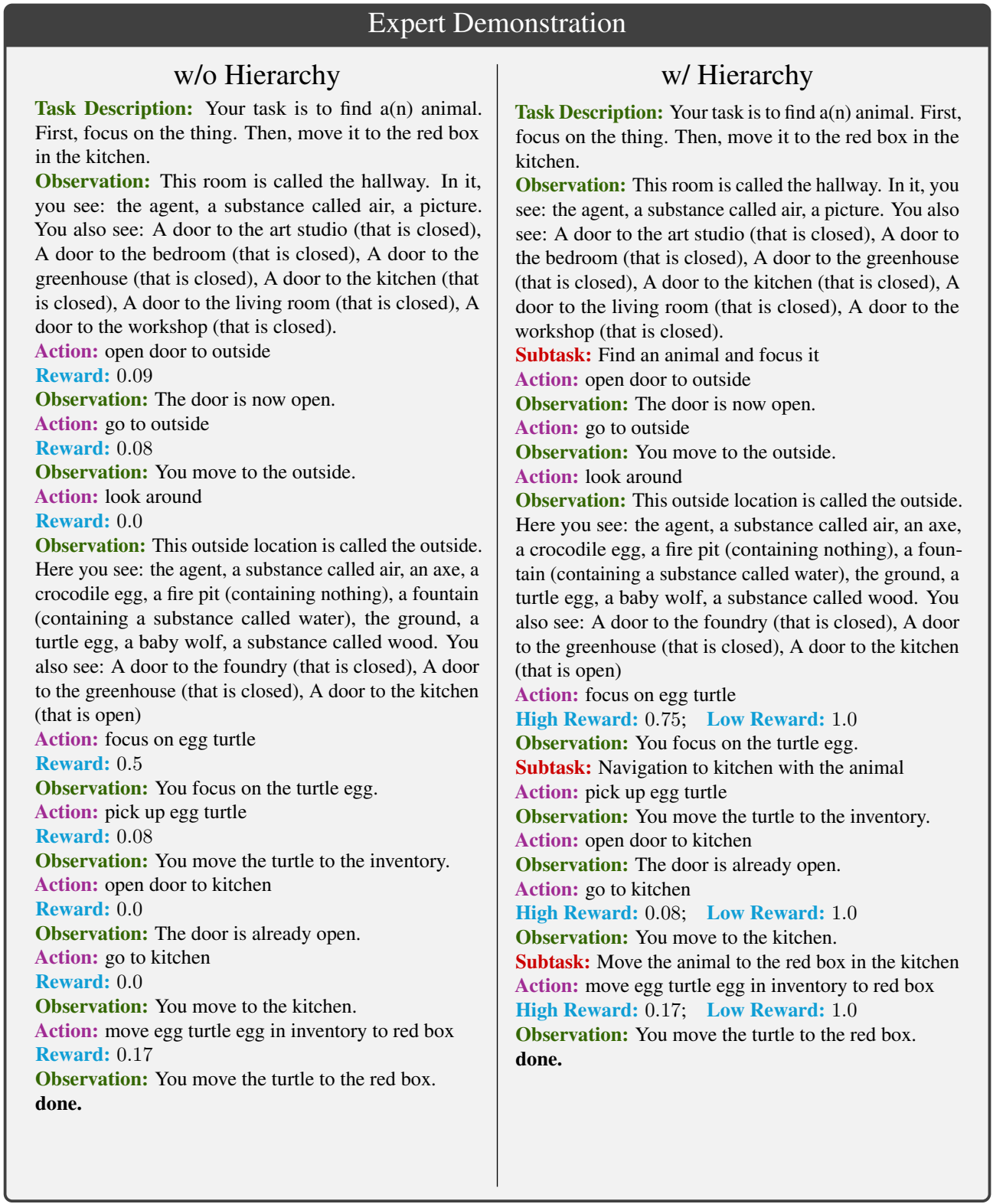

\centering
\scalebox{1.}{
\begin{tcolorbox}[center,breakable,title=\Large\centering{Expert Demonstration}]
\columnseprule=0.5pt
\begin{multicols}{2}

\begin{center}
{\Large{w/o Hierarchy}}
\end{center}
\textcolor{my_green}{\textbf{Task Description:}} Your task is to find a(n) animal. First, focus on the thing. Then, move it to the red box in the kitchen.\\
\textcolor{my_green}{\textbf{Observation:}} This room is called the hallway. In it, you see: the agent, a substance called air, a picture. You also see: A door to the art studio (that is closed), A door to the bedroom (that is closed), A door to the greenhouse (that is closed), A door to the kitchen (that is closed), A door to the living room (that is closed), A door to the workshop (that is closed).\\
\textcolor{my_purple}{\textbf{Action:}} open door to outside\\
\textcolor{my_blue}{\textbf{Reward:}} $0.09$\\
\textcolor{my_green}{\textbf{Observation:}} The door is now open.\\
\textcolor{my_purple}{\textbf{Action:}} go to outside\\
\textcolor{my_blue}{\textbf{Reward:}} $0.08$\\
\textcolor{my_green}{\textbf{Observation:}} You move to the outside.\\
\textcolor{my_purple}{\textbf{Action:}} look around\\
\textcolor{my_blue}{\textbf{Reward:}} $0.0$\\
\textcolor{my_green}{\textbf{Observation:}} This outside location is called the outside. Here you see: the agent, a substance called air, an axe, a crocodile egg, a fire pit (containing nothing), a fountain (containing a substance called water), the ground, a turtle egg, a baby wolf, a substance called wood. You also see: A door to the foundry (that is closed), A door to the greenhouse (that is closed), A door to the kitchen (that is open)\\
\textcolor{my_purple}{\textbf{Action:}} focus on egg turtle\\
\textcolor{my_blue}{\textbf{Reward:}} $0.5$\\
\textcolor{my_green}{\textbf{Observation:}} You focus on the turtle egg.\\
\textcolor{my_purple}{\textbf{Action:}} pick up egg turtle\\
\textcolor{my_blue}{\textbf{Reward:}} $0.08$\\
\textcolor{my_green}{\textbf{Observation:}} You move the turtle to the inventory.\\
\textcolor{my_purple}{\textbf{Action:}} open door to kitchen\\
\textcolor{my_blue}{\textbf{Reward:}} $0.0$\\
\textcolor{my_green}{\textbf{Observation:}} The door is already open.\\
\textcolor{my_purple}{\textbf{Action:}} go to kitchen\\
\textcolor{my_blue}{\textbf{Reward:}} $0.0$\\
\textcolor{my_green}{\textbf{Observation:}} You move to the kitchen.\\
\textcolor{my_purple}{\textbf{Action:}} move egg turtle egg in inventory to red box\\
\textcolor{my_blue}{\textbf{Reward:}} $0.17$\\
\textcolor{my_green}{\textbf{Observation:}} You move the turtle to the red box.\\
\textbf{done.}\\ \\ \\
\columnbreak

\begin{center}
{\Large {w/ Hierarchy}}
\end{center}
\textcolor{my_green}{\textbf{Task Description:}} Your task is to find a(n) animal. First, focus on the thing. Then, move it to the red box in the kitchen.\\
\textcolor{my_green}{\textbf{Observation:}} This room is called the hallway. In it, you see: the agent, a substance called air, a picture. You also see: A door to the art studio (that is closed), A door to the bedroom (that is closed), A door to the greenhouse (that is closed), A door to the kitchen (that is closed), A door to the living room (that is closed), A door to the workshop (that is closed).\\
\textcolor{my_red}{\textbf{Subtask:}} Find an animal and focus it\\
\textcolor{my_purple}{\textbf{Action:}} open door to outside\\
\textcolor{my_green}{\textbf{Observation:}} The door is now open.\\
\textcolor{my_purple}{\textbf{Action:}} go to outside\\
\textcolor{my_green}{\textbf{Observation:}} You move to the outside.\\
\textcolor{my_purple}{\textbf{Action:}} look around\\
\textcolor{my_green}{\textbf{Observation:}} This outside location is called the outside. Here you see: the agent, a substance called air, an axe, a crocodile egg, a fire pit (containing nothing), a fountain (containing a substance called water), the ground, a turtle egg, a baby wolf, a substance called wood. You also see: A door to the foundry (that is closed), A door to the greenhouse (that is closed), A door to the kitchen (that is open)\\
\textcolor{my_purple}{\textbf{Action:}} focus on egg turtle\\
\textcolor{my_blue}{\textbf{High Reward:}} $0.75$;\quad \textcolor{my_blue}{\textbf{Low Reward:}} $1.0$\\
\textcolor{my_green}{\textbf{Observation:}} You focus on the turtle egg.\\
\textcolor{my_red}{\textbf{Subtask:}} Navigation to kitchen with the animal\\
\textcolor{my_purple}{\textbf{Action:}} pick up egg turtle\\
\textcolor{my_green}{\textbf{Observation:}} You move the turtle to the inventory.\\
\textcolor{my_purple}{\textbf{Action:}} open door to kitchen\\
\textcolor{my_green}{\textbf{Observation:}} The door is already open.\\
\textcolor{my_purple}{\textbf{Action:}} go to kitchen\\
\textcolor{my_blue}{\textbf{High Reward:}} $0.08$;\quad \textcolor{my_blue}{\textbf{Low Reward:}} $1.0$\\
\textcolor{my_green}{\textbf{Observation:}} You move to the kitchen.\\
\textcolor{my_red}{\textbf{Subtask:}} Move the animal to the red box in the kitchen\\
\textcolor{my_purple}{\textbf{Action:}} move egg turtle egg in inventory to red box\\
\textcolor{my_blue}{\textbf{High Reward:}} $0.17$;\quad \textcolor{my_blue}{\textbf{Low Reward:}} $1.0$\\
\textcolor{my_green}{\textbf{Observation:}} You move the turtle to the red box.\\
\textbf{done.}
\end{multicols}
\end{tcolorbox}
}
\caption{
Expert demonstration for w/o and w/ hierarchical structure in ScienceWorld.
}
\label{fig:expert_traj}
\end{figure}

To support imitation learning, the ScienceWorld and ALFWorld provide human-annotated trajectories. For hierarchical data structuring, we utilize GPT-4 to decompose these trajectories into subtasks, creating a clear hierarchical representation of the demonstration data. We show an example expert demonstration trajectory for w/o and w/ hierarchical in ScienceWorld in Figure.~\ref{fig:expert_traj}

\subsection*{Medium Data Collection}
For medium-quality data collection, we employ two distinct sampling strategies:
\begin{enumerate}
\item \textbf{In-distribution Sampling:} During the SFT training process, we utilize the intermediate policy to sample sub-optimal trajectories on the training tasks.This sampling strategy helps better cover the task's world model, as the intermediate policy explores diverse solution paths and state transitions, leading to a more comprehensive understanding of the environment dynamics and task structure.
\item \textbf{Cross-task Generalization Sampling:} We partition the training tasks into $n$ (we set $n=3$) subsets. For each iteration, we train an SFT model on one subset and use it to collect data on the unseen training tasks from other subsets. This sampling strategy is specifically designed to capture trajectories that exhibit better generalization capabilities, as the policy is forced to adapt to novel but related tasks during data collection.
\end{enumerate}
Both of strategies approach ensures diversity in our medium-quality dataset, combining both task-specific expertise and cross-task generalization abilities. We provide examples of both strategies with and without hierarchical structure in Figure~\ref{fig:medium_traj_in_domain} and Figure~\ref{fig:medium_traj_out_domain}, respectively. 
\begin{figure}[!htp]
\centering
\scalebox{1.}{
\begin{tcolorbox}[center,breakable,title=\Large\centering{Medium Trajectory Collected by In-distribution Sampling Strategy}]
\columnseprule=0.5pt
\begin{multicols}{2}

\begin{center}
{\Large{w/o Hierarchy}}
\end{center}
\textcolor{my_green}{\textbf{Task Description:}} Your task is to boil apple juice. For compounds without a boiling point, combusting the substance is also acceptable. First, focus on the substance. Then, take actions that will cause it to change its state of matter.\\
\textcolor{my_green}{\textbf{Observation:}} This room is called the bathroom. In it, you see: the agent, substance called air, a bathtub, which is turned off. In the bathtub is: nothing. a glass cup (containing nothing), a picture, a sink, which is turned off. In the sink is: nothing. a toilet. In the toilet is: A drain, which is open, a substance called water. You also see: A door to the kitchen (that is closed).\\
\textcolor{my_purple}{\textbf{Action:}} open door to kitchen\\
\textcolor{my_blue}{\textbf{Reward:}} $0.0$\\
\textcolor{my_green}{\textbf{Observation:}} .\\
\textcolor{my_purple}{\textbf{Action:}} go to kitchen\\
\textcolor{my_blue}{\textbf{Reward:}} $0.03$\\
\textcolor{my_green}{\textbf{Observation:}} You move to the kitchen.\\
\textcolor{my_purple}{\textbf{Action:}} look around\\
\textcolor{my_blue}{\textbf{Reward:}} $0.0$\\
\textcolor{my_green}{\textbf{Observation:}} This room is called the kitchen. In it, you see: 
        a counter. On the counter is: a bowl (containing a red apple, a banana, an orange, a potato), a drawer.
        a cupboard. The cupboard door is closed. 
        a freezer. The freezer door is closed. 
        a fridge. The fridge door is closed. 
        a glass jar (containing a substance called sodium chloride)
        a lighter
        a oven, which is turned off. The oven door is closed. 
        a painting
        a sink, which is turned off. In the sink is: nothing.
        a substance called soap
        a stopwatch, which is deactivated. 
        a stove, which is turned off. On the stove is: nothing.
        a table. On the table is: a glass cup (containing nothing).
        a thermometer, currently reading a temperature of 10 degrees celsius
You also see:
        A door to the bathroom (that is open)
        A door to the hallway (that is closed)
        A door to the outside (that is closed).\\
\textcolor{my_purple}{\textbf{Action:}} pick up thermometer\\
\textcolor{my_blue}{\textbf{Reward:}} $0.0$\\
\textcolor{my_green}{\textbf{Observation:}} You move the thermometer to the inventory.\\
\textcolor{my_purple}{\textbf{Action:}} pick up metal pot\\
\textcolor{my_blue}{\textbf{Reward:}} $0.0$\\
\textcolor{my_green}{\textbf{Observation:}} You move the metal pot to the inventory.\\
\textcolor{my_purple}{\textbf{Action:}} pick up apple juice\\
\textcolor{my_blue}{\textbf{Reward:}} $0.0$\\
\textcolor{my_green}{\textbf{Observation:}} No known action matches that input.\\
\textbf{not done.}\\\\
\columnbreak

\begin{center}
{\Large {w/ Hierarchy}}
\end{center}
\textcolor{my_green}{\textbf{Task Description:}} Your task is to boil apple juice. For compounds without a boiling point, combusting the substance is also acceptable. First, focus on the substance. Then, take actions that will cause it to change its state of matter.\\
\textcolor{my_green}{\textbf{Observation:}} This room is called the bathroom. In it, you see: the agent, substance called air, a bathtub, which is turned off. In the bathtub is: nothing. a glass cup (containing nothing), a picture, a sink, which is turned off. In the sink is: nothing. a toilet. In the toilet is: A drain, which is open, a substance called water. You also see: A door to the kitchen (that is closed).\\
\textcolor{my_red}{\textbf{Subtask:}} Navigate to kitchen\\
\textcolor{my_purple}{\textbf{Action:}} open door to kitchen\\
\textcolor{my_green}{\textbf{Observation:}} The door is now open.\\
\textcolor{my_purple}{\textbf{Action:}} go to kitchen\\
\textcolor{my_blue}{\textbf{High Reward:}} $0.03$;\quad \textcolor{my_blue}{\textbf{Low Reward:}} $1.0$\\
\textcolor{my_green}{\textbf{Observation:}} You move to the kitchen.\\
\textcolor{my_red}{\textbf{Subtask:}} Prepare tools for measuring temperature and boiling\\
\textcolor{my_purple}{\textbf{Action:}} pick up thermometer\\
\textcolor{my_green}{\textbf{Observation:}}You move the thermometer to the inventory.\\
\textcolor{my_purple}{\textbf{Action:}} pick up metal pot\\
\textcolor{my_blue}{\textbf{High Reward:}} $0.0$;\quad \textcolor{my_blue}{\textbf{Low Reward:}} $1.0$\\
\textcolor{my_green}{\textbf{Observation:}} You move the metal pot to the inventory.\\
\textcolor{my_red}{\textbf{Subtask:}} Find apple juice and focus on it\\
\textcolor{my_purple}{\textbf{Action:}} open cupboard\\
\textcolor{my_green}{\textbf{Observation:}} The cupboard is now open.\\
\textcolor{my_purple}{\textbf{Action:}} focus on apple juice\\
\textcolor{my_blue}{\textbf{High Reward:}} $0.0$;\quad \textcolor{my_blue}{\textbf{Low Reward:}} $0.0$\\
\textcolor{my_green}{\textbf{Observation:}} No known action matches that input.\\
\textcolor{my_red}{\textbf{Subtask:}} Find apple juice and focus on it\\
\textcolor{my_purple}{\textbf{Action:}} open fridge\\
\textcolor{my_green}{\textbf{Observation:}} The fridge is now open.\\
\textcolor{my_purple}{\textbf{Action:}} focus on apple juice\\
\textcolor{my_blue}{\textbf{High Reward:}} $0.67$;\quad \textcolor{my_blue}{\textbf{Low Reward:}} $1.0$\\
\textcolor{my_green}{\textbf{Observation:}} You focus on the apple juice.\\
\textcolor{my_red}{\textbf{Subtask:}} Heat apple juice with stove\\
\textcolor{my_purple}{\textbf{Action:}} move cup containing apple juice to stove\\
\textcolor{my_green}{\textbf{Observation:}} You move the glass cup to the stove.\\
\textcolor{my_purple}{\textbf{Action:}} activate stove\\
\textcolor{my_blue}{\textbf{High Reward:}} $0.02$;\quad \textcolor{my_blue}{\textbf{Low Reward:}} $0.0$\\
\textcolor{my_green}{\textbf{Observation:}} The stove is broken.\\
\textcolor{my_red}{\textbf{Subtask:}} Can't heat apple juice because stove is broken\\
\textcolor{my_purple}{\textbf{Action:}} no action generated\\
\textcolor{my_blue}{\textbf{High Reward:}} $0.0$;\quad \textcolor{my_blue}{\textbf{Low Reward:}} $0.0$\\
\textcolor{my_green}{\textbf{Observation:}} No known action matches that input.\\
\textbf{not done.}
\end{multicols}
\end{tcolorbox}
}
\caption{
Comparison of w/o and w/ hierarchical trajectories collected through in-distribution sampling in ScienceWorld.
}
\label{fig:medium_traj_in_domain}
\end{figure}

\begin{figure}[!htp]
\centering
\scalebox{1.}{
\begin{tcolorbox}[center,breakable,title=\Large\centering{Medium Trajectory Collected by Cross-task Generalization Sampling Strategy}]
\columnseprule=0.5pt
\begin{multicols}{2}

\begin{center}
{\Large{w/o Hierarchy}}
\end{center}
\textcolor{my_green}{\textbf{Task Description:}} Your task is to boil apple juice. For compounds without a boiling point, combusting the substance is also acceptable. First, focus on the substance. Then, take actions that will cause it to change its state of matter.\\
\textcolor{my_green}{\textbf{Observation:}} This room is called the bathroom. In it, you see: the agent, substance called air, a bathtub, which is turned off. In the bathtub is: nothing. a glass cup (containing nothing), a picture, a sink, which is turned off. In the sink is: nothing. a toilet. In the toilet is: A drain, which is open, a substance called water. You also see: A door to the kitchen (that is closed).\\
\textcolor{my_purple}{\textbf{Action:}} open door to kitchen\\
\textcolor{my_blue}{\textbf{Reward:}} $0.0$\\
\textcolor{my_green}{\textbf{Observation:}} .\\
\textcolor{my_purple}{\textbf{Action:}} go to kitchen\\
\textcolor{my_blue}{\textbf{Reward:}} $0.03$\\
\textcolor{my_green}{\textbf{Observation:}} You move to the kitchen.\\
\textcolor{my_purple}{\textbf{Action:}} pick up thermometer\\
\textcolor{my_blue}{\textbf{Reward:}} $0.0$\\
\textcolor{my_green}{\textbf{Observation:}} You move the thermometer to the inventory.\\
\textcolor{my_purple}{\textbf{Action:}} pick up metal pot\\
\textcolor{my_blue}{\textbf{Reward:}} $0.0$\\
\textcolor{my_green}{\textbf{Observation:}} You move the metal pot to the inventory.\\
\textcolor{my_purple}{\textbf{Action:}} open fridge \\
\textcolor{my_blue}{\textbf{Reward:}} $0.0$\\
\textcolor{my_green}{\textbf{Observation:}} The fridge is now open.\\
\textcolor{my_purple}{\textbf{Action:}} focus on apple juice \\
\textcolor{my_blue}{\textbf{Reward:}} $0.67$\\
\textcolor{my_green}{\textbf{Observation:}} You focus on the apple juice.\\
\textcolor{my_purple}{\textbf{Action:}} move cup containing apple juice to stove\\
\textcolor{my_blue}{\textbf{Reward:}} $0.02$\\
\textcolor{my_green}{\textbf{Observation:}} You move the glass cup to the stove.\\
\textcolor{my_purple}{\textbf{Action:}} activate stove\\
\textcolor{my_blue}{\textbf{Reward:}} $0.0$\\
\textcolor{my_green}{\textbf{Observation:}} The stove is broken.\\
\textcolor{my_purple}{\textbf{Action:}} fix the stove\\
\textcolor{my_blue}{\textbf{Reward:}} $0.0$\\
\textcolor{my_green}{\textbf{Observation:}} No known action matches that input\\
\textbf{not done.}\\\\\\\\\\\\\\\\
\columnbreak

\begin{center}
{\Large {w/ Hierarchy}}
\end{center}
\textcolor{my_green}{\textbf{Task Description:}} Your task is to boil apple juice. For compounds without a boiling point, combusting the substance is also acceptable. First, focus on the substance. Then, take actions that will cause it to change its state of matter.\\
\textcolor{my_green}{\textbf{Observation:}} This room is called the bathroom. In it, you see: the agent, substance called air, a bathtub, which is turned off. In the bathtub is: nothing. a glass cup (containing nothing), a picture, a sink, which is turned off. In the sink is: nothing. a toilet. In the toilet is: A drain, which is open, a substance called water. You also see: A door to the kitchen (that is closed).\\
\textcolor{my_red}{\textbf{Subtask:}} Navigate to kitchen\\
\textcolor{my_purple}{\textbf{Action:}} open door to kitchen\\
\textcolor{my_green}{\textbf{Observation:}} The door is now open.\\
\textcolor{my_purple}{\textbf{Action:}} go to kitchen\\
\textcolor{my_blue}{\textbf{High Reward:}} $0.03$;\quad \textcolor{my_blue}{\textbf{Low Reward:}} $1.0$\\
\textcolor{my_green}{\textbf{Observation:}} You move to the kitchen.\\
\textcolor{my_red}{\textbf{Subtask:}} Prepare tools for measuring temperature and boiling\\
\textcolor{my_purple}{\textbf{Action:}} pick up thermometer\\
\textcolor{my_green}{\textbf{Observation:}}You move the thermometer to the inventory.\\
\textcolor{my_purple}{\textbf{Action:}} pick up metal pot\\
\textcolor{my_blue}{\textbf{High Reward:}} $0.0$;\quad \textcolor{my_blue}{\textbf{Low Reward:}} $1.0$\\
\textcolor{my_green}{\textbf{Observation:}} You move the metal pot to the inventory.\\
\textcolor{my_red}{\textbf{Subtask:}} Find apple juice and focus on it\\
\textcolor{my_purple}{\textbf{Action:}} open cupboard\\
\textcolor{my_green}{\textbf{Observation:}} The cupboard is now open.\\
\textcolor{my_purple}{\textbf{Action:}} focus on apple juice\\
\textcolor{my_blue}{\textbf{High Reward:}} $0.0$;\quad \textcolor{my_blue}{\textbf{Low Reward:}} $0.0$\\
\textcolor{my_green}{\textbf{Observation:}} No known action matches that input.\\
\textcolor{my_red}{\textbf{Subtask:}} Find apple juice and focus on it\\
\textcolor{my_purple}{\textbf{Action:}} open fridge\\
\textcolor{my_green}{\textbf{Observation:}} The fridge is now open.\\
\textcolor{my_purple}{\textbf{Action:}} focus on apple juice\\
\textcolor{my_blue}{\textbf{High Reward:}} $0.67$;\quad \textcolor{my_blue}{\textbf{Low Reward:}} $1.0$\\
\textcolor{my_green}{\textbf{Observation:}} You focus on the apple juice.\\
\textcolor{my_red}{\textbf{Subtask:}} Heat apple juice with stove\\
\textcolor{my_purple}{\textbf{Action:}} move cup containing apple juice to stove\\
\textcolor{my_green}{\textbf{Observation:}} You move the glass cup to the stove.\\
\textcolor{my_purple}{\textbf{Action:}} activate stove\\
\textcolor{my_blue}{\textbf{High Reward:}} $0.0$;\quad \textcolor{my_blue}{\textbf{Low Reward:}} $0.0$\\
\textcolor{my_green}{\textbf{Observation:}} The stove is broken.\\
\textcolor{my_red}{\textbf{Subtask:}} Find other stove to heat apple juice\\
\textcolor{my_purple}{\textbf{Action:}} focus on stove\\
\textcolor{my_blue}{\textbf{High Reward:}} $0.0$;\quad \textcolor{my_blue}{\textbf{Low Reward:}} $0.0$\\
\textcolor{my_green}{\textbf{Observation:}} No known action matches that input.\\
\textbf{not done.}
\end{multicols}
\end{tcolorbox}
}
\caption{
Comparison of w/o and w/ hierarchical trajectories collected by cross-task generalization sampling in ScienceWorld.
}
\label{fig:medium_traj_out_domain}
\end{figure}

In Figure~\ref{fig:medium_traj_in_domain} (in-distribution sampling), we can see the  the exploration without hierarchy terminates early after attempting to \texttt{"pick up apple juice"}. In contrast, the hierarchical approach extends further through clear subtask planning (\texttt{"Prepare tools"}, \texttt{"Find apple juice"}, \texttt{"Heat apple juice"}), continuing until it discovers the stove malfunction.

In Figure~\ref{fig:medium_traj_out_domain} (cross-task generalization sampling) demonstrates enhanced exploration capabilities. Under the same initial conditions, the non-hierarchical approach attempts to \texttt{"fix the stove"}, while the hierarchical approach not only identifies the stove malfunction but actively seeks alternative solutions (\texttt{"Find other stove to heat apple juice"}), showing more sophisticated problem-solving strategies.

\subsection*{Data Structure and Setups}
The high-level and low-level training dataset structure follows a sequential format that captures the complete interaction trajectory:
\begin{tcolorbox}[breakable,center,breakable,title=\Large\centering{Training Dataset Structure}]\label{box:traj}
\textcolor{my_green}{\textbf{High-Level Trajectory:}}\\
\{\textcolor{my_red}{high prompt}, \textcolor{my_red}{task description}, \textcolor{my_blue}{obs 0}, \textcolor{my_purple}{subtask 0}, \textcolor{my_orange}{high reward 0} $...$ \textcolor{my_blue}{obs T-1}, \textcolor{my_purple}{subtask T-1}, \textcolor{my_orange}{high reward T-1}, \textcolor{my_blue}{obs T}\}
\\\\
\textcolor{my_green}{\textbf{Low-Level Trajectory:}}\\
\{\textcolor{my_red}{Low prompt}, \textcolor{my_red}{subtask 0}, \textcolor{my_blue}{obs 0}, \textcolor{my_purple}{action 0}, \textcolor{my_orange}{low reward 1} $...$ \textcolor{my_blue}{obs c-1}, \textcolor{my_purple}{action c-1}, \textcolor{my_orange}{low reward c-1}, \textcolor{my_blue}{obs c}\}
\end{tcolorbox}

\subsection*{O2O task Setups}
To evaluate the generalization capabilities of our method, we construct an O2O (Online-to-Offline) dataset covering three distinct domains: electrical, biology, and thermodynamics. Each domain contains one test task and multiple train tasks, as shown in Table~\ref{tab:online_task}. For the electrical domain, we have \texttt{"test-conductivity"} as the test task, along with three train tasks related to conductivity testing and power components. The biology domain features \texttt{"find-animal"} as the test task, accompanied by ten train tasks involving various biological concepts such as living/non-living identification, plant growth, and lifespan studies. In the thermodynamics domain, \texttt{"boil"} serves as the test task, supported by seven train tasks covering different aspects of state changes and chemical mixing processes. This setup provides a rigorous test of the agent's ability to transfer knowledge from trained tasks to novel but related domains in the ScienceWorld environment.

\begin{table}[htbp]
\centering
\setlength{\tabcolsep}{15pt}  
\caption{Distribution of test and train tasks across electrical, biology, and thermodynamics domains}
\begin{tabular}{@{}c|c|c} 
\toprule
\textbf{Domain }& \textbf{Task Name} & \textbf{Type} \\
\midrule
\multirow{4}{*}{\parbox{2.5cm}{\centering Electrical}} 
& \cellcolor{gray!20}test-conductivity & \cellcolor{gray!20}Test \\
& test-conductivity-of-unknown-substances & Train \\
& power-component & Train \\
& power-component-renewable-vs-nonrenewable-energy & Train \\
\midrule
\multirow{11}{*}{\parbox{2.5cm}{\centering Biology}} 
& \cellcolor{gray!20}find-animal & \cellcolor{gray!20}Test \\
& find-living-thing & Train \\
& find-non-living-thing & Train \\
& find-plant & Train \\
& grow-fruit & Train \\
& grow-plant & Train \\
& identify-life-stages-1 & Train \\
& identify-life-stages-2 & Train \\
& lifespan-longest-lived & Train \\
& lifespan-longest-lived-then-shortest-lived & Train \\
& lifespan-shortest-lived & Train \\
\midrule
\multirow{8}{*}{\parbox{2.5cm}{\centering Thermodynamics}} 
& \cellcolor{gray!20}boil & \cellcolor{gray!20}Test \\
& freeze & Train \\
& melt & Train \\
& change-the-state-of-matter-of & Train \\
& chemistry-mix & Train \\
& chemistry-mix-paint-secondary-color & Train \\
& chemistry-mix-paint-tertiary-color & Train \\
& use-thermometer & Train \\
\bottomrule
\end{tabular}
\label{tab:online_task}
\end{table}

\subsection*{Reward Design}
\begin{itemize}
\item \textbf{ScienceWorld}: The agent receives a dense reward ranging from 0 to 1 at each step, reflecting the continuous progress in scientific experimentation tasks across 30 scenarios in 10 categories.
\item \textbf{ALFWorld}: The agent receives a sparse binary reward (0 or 1), where 1 is given only upon successful completion of household navigation and manipulation tasks, and 0 otherwise.
\end{itemize}

\begin{itemize}
    \item \textbf{High-level Reward}: The high-level policy accumulates environmental rewards upon the completion of subtasks by the low-level policy, reflecting the agent's progress in achieving the overall objective.
    \item \textbf{Low-level Reward}: The low-level policy receives binary rewards from the high-level policy, indicating whether a subtask is successfully completed or not.
\end{itemize}

\clearpage
\section*{Appendix C. Training and Evaluation setups}
\subsection*{Models}
We build our method on three open-source language models:
1) \texttt{Mistral-7B} \citep{jiang2023mistral}, the Mistral-7B-Instruct-v0.2 version. 
2) \texttt{Gemma-7B} \citep{team2024gemma}, the Gemma-1.1-7B-it version. 
3) \texttt{Llama-3-8B} \citep{llama3}, the Meta-Llama-3-8B-Instruct version.
We employ LoRA for parameter-efficient fine-tuning of all language models.

\subsection*{Baselines}
% We compare GLIDER against several strong baselines:
We compare GLIDER against various strong baselines:
% 1) \texttt{ReAct}~\citep{yao2023react}, A pioneering approach that incorporates Chain-of-Thought (CoT) prompting in decision-making tasks through a structured Thought-Action-Observation loop.
1) \texttt{ReAct}~\citep{yao2023react}, a pioneering approach that incorporates CoT prompting in decision-making tasks through a structured Thought-Action-Observation loop.
2) \texttt{Reflexion}~\citep{shinn2023reflexion}, an advanced prompt-based framework that enhances agent decision-making through self-reflective verbal feedback.
3) \texttt{SwiftSage}~\citep{lin2023swiftsage}, a dual-process theory-based cognitive framework that integrates the strengths of behavior cloning and prompting for complex interactive reasoning and action-planning tasks.
4) \texttt{NAT}~\citep{wang2024learning}, a fine-tuning approach that enables LLMs to learn from failure trajectories through quality control and fine-tuning strategies.
5) \texttt{ETO}~\citep{song2024trial}, an iterative optimization framework between exploring the environment to collect contrastive trajectory pairs and fine-tuning the LLM policy using DPO~\citep{rafailov2023direct}.

\subsection*{Hyperparameter}
we employ LoRA for parameter-efficient fine-tuning for all language models. During the SFT phase, we train for $5$ epochs with a batch size of $32$, using the AdamW optimizer with a learning rate of $1e-4$. The detailed hyperparameters are shown in the following table~\ref{tab:para}.
Policy are evaluated on both seen and unseen tasks across two benchmarks.
For the ORL phase, we train for 4 epochs with different learning rates for the actor ($1e-5$) and critic ($1e-4$) networks. The target critic network is updated using soft updates with $\tau = 0.2$, and we set the advantage weighted factor $\lambda$ to $0.99$. The training data consists of expert and medium data in a 1:2 ratio.
\begin{table}[h] \setlength{\tabcolsep}{2mm}
\caption{Hyperparameters used for GLIDER.}
\label{tab:para}
\begin{center}
\renewcommand{\arraystretch}{1.3}
\begin{tabular}{lrlr}
\hline 
\toprule[0.5pt]
\multicolumn{1}{l}{\bf Hyperparameter}  &\multicolumn{1}{l}{\bf Value} &
\multicolumn{1}{l}{\bf Hyperparameter}  &\multicolumn{1}{l}{\bf Value} \\
\hline 
batch size                  &64                     &temperature                                &0.7\\
batch size per device       &2                      &advantage weighted factor $\lambda$        &0.99\\
gradient accumulation steps &8                      &soft update $\tau$                         &0.2 \\
actor learning rate         &$1\times 10^{-5}$      &discount factor $\gamma$                   &0.99\\
critic learning rate        &$1\times 10^{-4}$      &sft epochs                                 &5\\
sft learning rate           &$1\times 10^{-4}$      &orl epochs                                 &4\\
lora r                      &16                     &data mixture radio                         &1:2\\
lora alpha                  &32                     &warmup ratio                               &0.03\\
lora dropout                &0.05                   &max new tokens                             &32    \\
\hline 
\toprule[0.5pt]
\end{tabular}
\end{center}
\end{table}
% In the O2O phase, we assess the generalization performance by evaluating on 3 held-out tasks from ScienceWorld that were not used during training. We track the evaluation metrics during the online adaptation process and compare our hierarchical approach with several baselines, including standard Actor-Critic, AWAC, and Actor-Critic hierarchical variant. Please refer to appendix C for detailed training and evaluation setups, hyperparameters, and prompt templates for both high-level and low-level inputs.

\clearpage
\section*{Appendix D. Prompts and Case Study}
We illustrate the prompts used in our paper, where High-Level and Low-Level prompts establish hierarchical control between different agents, while Check Subtask Complete Prompt enables the high-level agent to evaluate subtask completion by the low-level agent.
\begin{tcolorbox}[breakable,title=\Large\centering{Prompts}]\label{fig:prompt}
\textcolor{my_green}{\textbf{High-Level Prompt:}}\\
You are a high-level planner. Based on the state (task description, group action and current observation), please generate a clear and simple subtask.
\\\\
\textcolor{my_green}{\textbf{Low-Level Prompt:}}\\
You are a low-level action executor. Based on the current subtask and observation, please generate a executable action and determine if the subtask is completed (true/false).
\\\\
\textcolor{my_green}{\textbf{Check Subtask Complete Prompt:}}\\
{Determine if the low-level actions successfully completed the given subtask by high-level:\\
Subtask: $[$subtask$]$\\
Initial observation: $[$initial obs$]$\\
Actions: $[$action sequence$]$\\
Final observation: $[$final obs $]$\\
Output only a single digit:\\
True if the actions successfully completed the subtask\\
False if the actions failed to complete the subtask\\
Give the "True" or "False":}
\end{tcolorbox}

\begin{figure}[!htp]
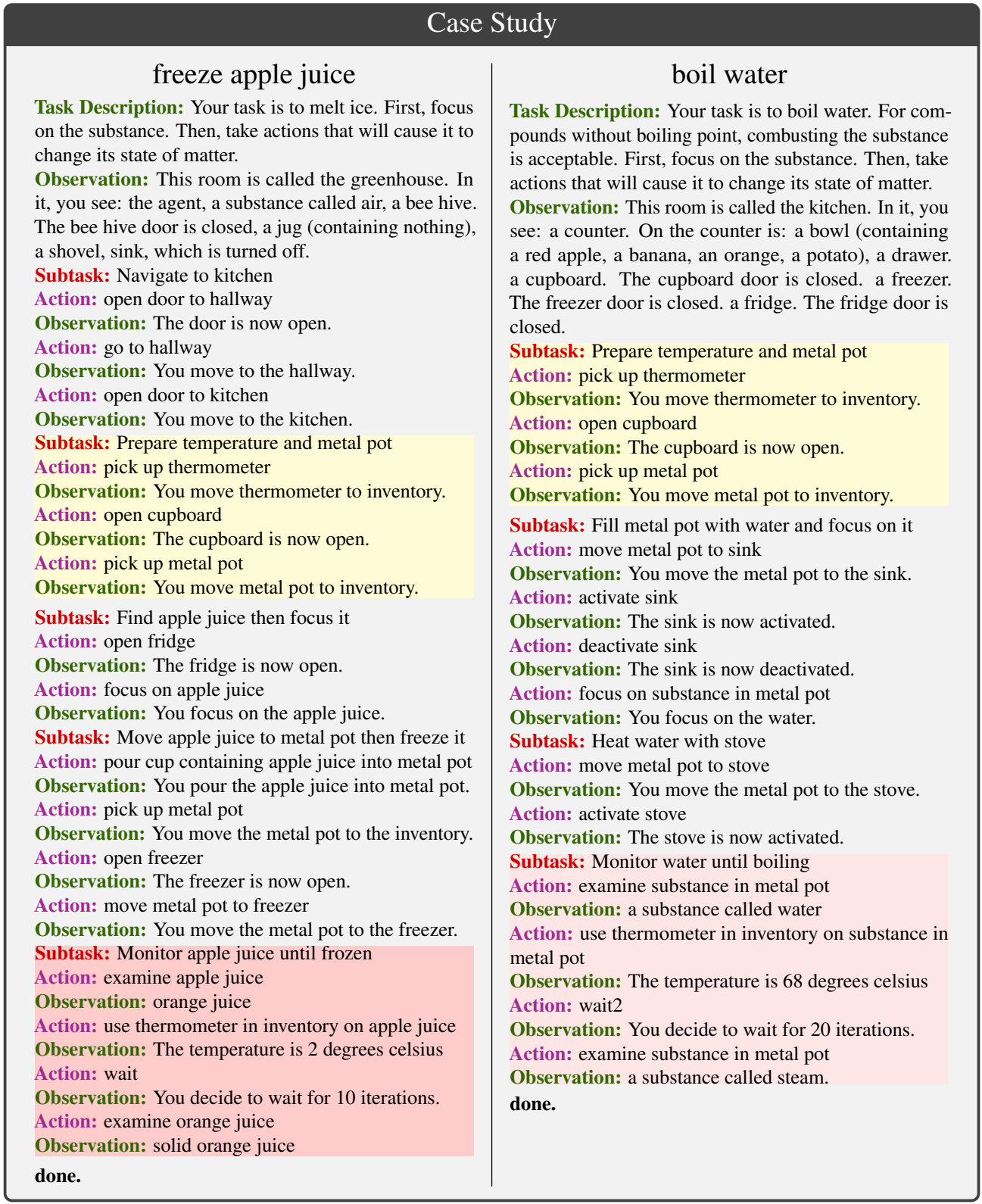

\centering
\scalebox{1.}{
\begin{tcolorbox}[center,breakable,title=\Large\centering{Case Study}]
\columnseprule=0.5pt
\begin{multicols}{2}

\begin{center}
{\Large{freeze apple juice}}
\end{center}
\textcolor{my_green}{\textbf{Task Description:}} Your task is to melt ice. First, focus on the substance. Then, take actions that will cause it to change its state of matter.\\
\textcolor{my_green}{\textbf{Observation:}} This room is called the greenhouse. In it, you see: the agent, a substance called air, a bee hive. The bee hive door is closed, a jug (containing nothing), a shovel, sink, which is turned off.\\
\textcolor{my_red}{\textbf{Subtask:}} Navigate to kitchen\\
\textcolor{my_purple}{\textbf{Action:}} open door to hallway\\
\textcolor{my_green}{\textbf{Observation:}} The door is now open.\\
\textcolor{my_purple}{\textbf{Action:}} go to hallway\\
\textcolor{my_green}{\textbf{Observation:}} You move to the hallway.\\
\textcolor{my_purple}{\textbf{Action:}} open door to kitchen\\
\textcolor{my_green}{\textbf{Observation:}} You move to the kitchen.\\
{\setlength{\fboxsep}{0pt}%    % 设置内边距为0
\setlength{\fboxrule}{0pt}%    % 设置边框宽度为0
\colorbox{yellow!20}{%
\parbox[t]{\dimexpr\columnwidth}{%  % 使用\dimexpr确保宽度精确
\textcolor{my_red}{\textbf{Subtask:}} Prepare temperature and metal pot\\
\textcolor{my_purple}{\textbf{Action:}} pick up thermometer\\
\textcolor{my_green}{\textbf{Observation:}} You move thermometer to inventory.\\
\textcolor{my_purple}{\textbf{Action:}} open cupboard\\
\textcolor{my_green}{\textbf{Observation:}} The cupboard is now open.\\
\textcolor{my_purple}{\textbf{Action:}} pick up metal pot\\
\textcolor{my_green}{\textbf{Observation:}} You move metal pot to inventory.
}}}\\\\
\textcolor{my_red}{\textbf{Subtask:}} Find apple juice then focus it\\
\textcolor{my_purple}{\textbf{Action:}} open fridge\\
\textcolor{my_green}{\textbf{Observation:}} The fridge is now open.\\
\textcolor{my_purple}{\textbf{Action:}} focus on apple juice\\
\textcolor{my_green}{\textbf{Observation:}} You focus on the apple juice.\\
\textcolor{my_red}{\textbf{Subtask:}} Move apple juice to metal pot then freeze it\\
\textcolor{my_purple}{\textbf{Action:}} pour cup containing apple juice into metal pot\\
\textcolor{my_green}{\textbf{Observation:}} You pour the apple juice into metal pot.\\
\textcolor{my_purple}{\textbf{Action:}} pick up metal pot\\
\textcolor{my_green}{\textbf{Observation:}} You move the metal pot to the inventory.\\
\textcolor{my_purple}{\textbf{Action:}} open freezer\\
\textcolor{my_green}{\textbf{Observation:}} The freezer is now open.\\
\textcolor{my_purple}{\textbf{Action:}} move metal pot to freezer\\
\textcolor{my_green}{\textbf{Observation:}} You move the metal pot to the freezer.\\
{\setlength{\fboxsep}{0pt}%    % 设置内边距为0
\setlength{\fboxrule}{0pt}%    % 设置边框宽度为0
\colorbox{red!20}{%
\parbox[t]{\dimexpr\columnwidth}{%  % 使用\dimexpr确保宽度精确
\textcolor{my_red}{\textbf{Subtask:}} Monitor apple juice until frozen\\
\textcolor{my_purple}{\textbf{Action:}} examine apple juice\\
\textcolor{my_green}{\textbf{Observation:}} orange juice\\
\textcolor{my_purple}{\textbf{Action:}} use thermometer in inventory on apple juice\\
\textcolor{my_green}{\textbf{Observation:}} The temperature is 2 degrees celsius\\
\textcolor{my_purple}{\textbf{Action:}} wait\\
\textcolor{my_green}{\textbf{Observation:}} You decide to wait for 10 iterations.\\
\textcolor{my_purple}{\textbf{Action:}} examine orange juice\\
\textcolor{my_green}{\textbf{Observation:}} solid orange juice
}}}\\\\
\textbf{done.} 
\columnbreak

\begin{center}
{\Large {boil water}}
\end{center}
\textcolor{my_green}{\textbf{Task Description:}} Your task is to boil water. For compounds without boiling point, combusting the substance is acceptable. First, focus on the substance. Then, take actions that will cause it to change its state of matter.\\
\textcolor{my_green}{\textbf{Observation:}} This room is called the kitchen. In it, you see: 
        a counter. On the counter is: a bowl (containing a red apple, a banana, an orange, a potato), a drawer.
        a cupboard. The cupboard door is closed. 
        a freezer. The freezer door is closed. 
        a fridge. The fridge door is closed. \\
{\setlength{\fboxsep}{0pt}%    % 设置内边距为0
\setlength{\fboxrule}{0pt}%    % 设置边框宽度为0
\colorbox{yellow!20}{%
\parbox[t]{\dimexpr\columnwidth}{%  % 使用\dimexpr确保宽度精确
\textcolor{my_red}{\textbf{Subtask:}} Prepare temperature and metal pot\\
\textcolor{my_purple}{\textbf{Action:}} pick up thermometer\\
\textcolor{my_green}{\textbf{Observation:}} You move thermometer to inventory.\\
\textcolor{my_purple}{\textbf{Action:}} open cupboard\\
\textcolor{my_green}{\textbf{Observation:}} The cupboard is now open.\\
\textcolor{my_purple}{\textbf{Action:}} pick up metal pot\\
\textcolor{my_green}{\textbf{Observation:}} You move metal pot to inventory.
}}}\\\\
\textcolor{my_red}{\textbf{Subtask:}} Fill metal pot with water and focus on it\\
\textcolor{my_purple}{\textbf{Action:}} move metal pot to sink\\
\textcolor{my_green}{\textbf{Observation:}} You move the metal pot to the sink.\\
\textcolor{my_purple}{\textbf{Action:}} activate sink\\
\textcolor{my_green}{\textbf{Observation:}} The sink is now activated.\\
\textcolor{my_purple}{\textbf{Action:}} deactivate sink \\
\textcolor{my_green}{\textbf{Observation:}} The sink is now deactivated.\\
\textcolor{my_purple}{\textbf{Action:}} focus on substance in metal pot\\
\textcolor{my_green}{\textbf{Observation:}} You focus on the water.\\
\textcolor{my_red}{\textbf{Subtask:}} Heat water with stove\\
\textcolor{my_purple}{\textbf{Action:}} move metal pot to stove\\
\textcolor{my_green}{\textbf{Observation:}} You move the metal pot to the stove.\\
\textcolor{my_purple}{\textbf{Action:}} activate stove\\
\textcolor{my_green}{\textbf{Observation:}} The stove is now activated.\\
{\setlength{\fboxsep}{0pt}%    % 设置内边距为0
\setlength{\fboxrule}{0pt}%    % 设置边框宽度为0
\colorbox{red!10}{%
\parbox[t]{\dimexpr\columnwidth}{%  % 使用\dimexpr确保宽度精确
\textcolor{my_red}{\textbf{Subtask:}} Monitor water until boiling\\
\textcolor{my_purple}{\textbf{Action:}} examine substance in metal pot\\
\textcolor{my_green}{\textbf{Observation:}} a substance called water\\
\textcolor{my_purple}{\textbf{Action:}} use thermometer in inventory on substance in metal pot\\
\textcolor{my_green}{\textbf{Observation:}} The temperature is 68 degrees celsius\\
\textcolor{my_purple}{\textbf{Action:}} wait2\\
\textcolor{my_green}{\textbf{Observation:}} You decide to wait for 20 iterations.\\
\textcolor{my_purple}{\textbf{Action:}} examine substance in metal pot\\
\textcolor{my_green}{\textbf{Observation:}} a substance called steam.
}}}\\\\
\textbf{done.}
\end{multicols}
\end{tcolorbox}
}
\caption{
Hierarchical decomposition reveals shared subtask patterns across two example tasks in ScienceWorld.
}
\label{fig:case}
\end{figure}

To demonstrate the advantages of hierarchical decomposition, we present a case study show in Figure.~\ref{fig:case} comparing two different tasks: freezing apple juice and boiling water.
The highlighted portions reveal two types of structural similarities: identical subtasks (highlighted in yellow) that can be directly reused, and analogous subtasks (highlighted in red) that share similar underlying patterns despite different objectives. For example, both tasks contain the identical subtask \texttt{"Prepare temperature and metal pot"}, while \texttt{"Monitor apple juice until frozen"} and \texttt{"Monitor water until boiling"} represent analogous patterns of state monitoring and waiting. This hierarchical approach not only enables direct subtask reuse, but also allows the planner to recognize and adapt similar strategic patterns across different tasks, thereby reducing the planning complexity and improving efficiency.

\end{document}